\documentclass[preprint,12pt]{elsarticle}
\usepackage[margin=2.5cm]{geometry}
\usepackage{xcolor}
\usepackage{booktabs}
\biboptions{sort&compress}
\usepackage{setspace}
\usepackage{placeins}

\usepackage{amssymb}

\journal{Additive Manufacturing Letters}

\begin{document}

\begin{frontmatter}


\title{Deep Learning for Melt Pool Depth Contour Prediction From Surface Thermal Images via Vision Transformers}

\author[inst1]{Francis Ogoke}

\affiliation[inst1]{organization={Department of Mechanical Engineering},
            addressline={Carnegie Mellon University}, 
            city={Pittsburgh},
            postcode={15213}, 
            state={PA},
            country={USA}}

\author[inst1]{Peter Pak}
\author[inst1]{Alexander Myers}
\author[inst1]{Guadalupe Quirarte}
\author[inst1]{Jack Beuth}
\author[inst1]{Jonathan Malen}

\author[inst1,inst2,inst3]{Amir Barati Farimani\corref{cor1}}
\cortext[cor1]{Corresponding author, barati@cmu.edu}

\affiliation[inst2]{organization={Department of Chemical Engineering, Carnegie Mellon University},
            city={Pittsburgh},
            postcode={15213}, 
            state={PA},
            country={USA}}
            
\affiliation[inst3]{organization={Machine Learning Department, Carnegie Mellon University},
            city={Pittsburgh},
            postcode={15213}, 
            state={PA},
            country={USA}}
\begin{abstract}
Insufficient melt depths during Laser Powder Bed Fusion (L-PBF) can lead to lack-of-fusion defects and deteriorated mechanical and fatigue performance. In-situ monitoring of the melt pool subsurface morphology requires specialized equipment that may not be readily accessible or scalable. Therefore, we introduce a machine learning framework to correlate in-situ two-color thermal images observed via high-speed color imaging to the two-dimensional profile of the melt pool cross-section. Specifically, we employ a hybrid CNN-Transformer architecture to establish a correlation between single bead off-axis thermal image sequences and melt pool cross-section contours measured via optical microscopy. In this architecture, a ResNet model embeds the spatial information contained within the thermal images to a latent vector, while a Transformer model correlates the sequence of embedded vectors to extract temporal information.  The performance of this model is evaluated through dimensional and geometric comparisons to the corresponding experimental melt pool observations. Our framework is able to model the curvature of the subsurface melt pool structure, with improved performance in high energy density regimes compared to analytical melt pool models. Additionally, we note that the use of ratiometric temperature estimates improves the accuracy of the model predictions compared to monochromatic imaging.
\end{abstract}



\end{frontmatter}


\section*{Introduction}
\label{sec:introduction}

 Laser Powder Bed Fusion (L-PBF) is a metal additive manufacturing process that decreases material waste and increases local control when compared to conventional manufacturing processes \cite{li2020review, reeves2011additive}. Due to these advantages, L-PBF has become a useful production technique for industrial applications requiring parts with complex internal structure  \cite{cunningham2017analyzing,mower2016mechanical,spierings2013fatigue,lewandowski2016metal, tofail2018additive}. During L-PBF, a product is iteratively built by fusing successive layers of metal alloy powder together, using a laser heat source to induce melting at a specific cross-section \cite{yadroitsev2015hierarchical}. However, due to the process complexity and short length and time scales involved, the stochastic formation of defects can increase performance variability and limit adoption for use cases with strict part specifications \cite{zhang2018evolution, khairallah2016laser}. For instance, the transient behavior of the instantaneous melt pool directly changes the residual stress, microstructure, and porosity within the part \cite{matthews2020controlling, li2018residual, koepf20183d}. At high energy density, unstable vapor cavities within the part can collapse and cause spherical gas voids, while at low energy density, insufficient melting between layers results in lack-of-fusion porosity \cite{gong2014analysis, ning2019analytical}. These defects have a significant effect on the mechanical and fatigue performance of the finalized parts, localizing stress and reducing the life cycle until fatigue induced failure \cite{ronneberg2020revealing, rice1997limitations, wilson2021combined}.

The formation of porosity during L-PBF is intrinsically linked to the morphology of the melt pool during printing \cite{bayat2019keyhole, tang2017prediction}. Therefore, several methods for estimating the morphology of the melt pool during printing have been proposed. Analytical methods based on simplified models of the heat transfer phenomena during L-PBF provide rapid estimates of the melt pool size, but lose accuracy in high energy density regimes of the melting process \cite{tang2017prediction, rosenthal1941mathematical, eagar1983temperature}. Multiphysics simulations produce more accurate models of the melt pool behavior during printing, accounting for phase change, vaporization, and variable absorptivity effects \cite{cheng2019computational, khairallah2016laser, markl2016multiscale}. However, these simulations are based upon nominal specifications of the process conditions, and do not generalize to cases where unexpected phenomena lead to variations in the input conditions \cite{gaikwad2022multi}

Due to the challenges involved in estimating the morphology of the melt pool with both simulation-based and experimental approaches, recent work has leveraged machine learning with experimental \textit{in-situ} monitoring and simulation modeling to create estimates of the defect formation behavior during printing \cite{Tian2020Deep, taherkhani2022unsupervised, strayer2022accelerating, hemmasian2023surrogate, ogoke2023inexpensive}. Machine learning techniques have shown promise in navigating the high-dimensional, uncertain search spaces within scientific and engineering applications \cite{jadhav2023stressd, ogoke2021thermal, chen2022data}. Therefore, these techniques are also readily applicable to the task of correlating in-situ monitoring information to post-build, ex-situ characterization.

Early approaches to machine learning enabled process monitoring were based on layerwise optical monitoring. In-situ optical monitoring enables process monitoring of the powder bed at distance from the build plate, providing information on the current state of the powder bed surface \cite{buchbinder2011high, gobert2018application, scime2018anomaly}.  Gobert et al. demonstrate the use of a high resolution digital single-lens reflex camera to correlate layerwise images of the process to the presence of defects within X-ray CT data \cite{gobert2018application}. In related work, Scime et al. demonstrate the use of bag-of-words clustering to correlate filter responses applied to optical images to potential defect classes \cite{scime2018anomaly}. However, it can be difficult to resolve temporal process events to these optical images, as they are typically observed following the completion of an entire layer \cite{shevchik2018acoustic}. Additionally, the large scale of this monitoring system may not always resolve small, localized pore defects \cite{scime2019melt}.

In response, researchers have developed small-scale melt pool imaging tools, enabling more direct linkage to the melt pool dimensions and defect regimes. For instance, Gaikwad et al. monitored the melt pool with coaxial high-speed video cameras at two different wavelengths \cite{gaikwad2022multi}. From this data, they designed features to link the sensor information to the type of porosity present. In an alternate work, Larsen et al. present FlawNet, a method to detect anomalous melt pool instances potentially indicative of porosity through autoregressive training on the temporal dynamics of the melting process \cite{larsen2022deep}. While these works show promise in detecting porosity within the finalized part, these models are built upon binary classification algorithms, and do not directly characterize the degree of porosity present in the material, or account for potential pore formation mechanisms.

Therefore, we propose an approach that links experimental, high-speed \textit{in-situ} videos of the melt pool to the \textit{ex-situ} melt pool morphology beneath the surface. Specifically, high-speed off-axis image sequences of the melt pool surface temperature observed during printing are collected and mapped to the corresponding solidified melt pool cross-section. The ability to predict the melt pool contour from surface images provides a pathway towards \textit{in-situ} control for defect mitigation. Our model employs a transformer-based architecture, originally developed for sequence modeling in natural language processing tasks \cite{vaswani2017attention}, to analyze these image sequences. The transformer architecture uses attention mechanisms to learn how elements in a sequence affect each other and inform downstream prediction tasks. To reduce the complexity of our model, we decompose the architecture into spatial and temporal components. Specifically, we leverage a convolutional encoder to model the spatial information, and the transformer model to extract temporal information. With this framework, we also examine the influence of transfer learning from simulation data to the experimental domain to reduce the amount of manual data collection required. The performance of the model is evaluated by comparing the dimensions of the predicted cross-sectional areas the shape of the predicted contours, and the Intersection-over-Union (IoU) score of the contour reconstructions. Finally, we validate our model through comparisons to \textit{ex-situ} characterization of multi-track prints.

\section*{Methodology}
\label{sec:methods}

\label{subsec:mlarch}

\begin{figure}[hbt!]

\includegraphics[width=1\linewidth]{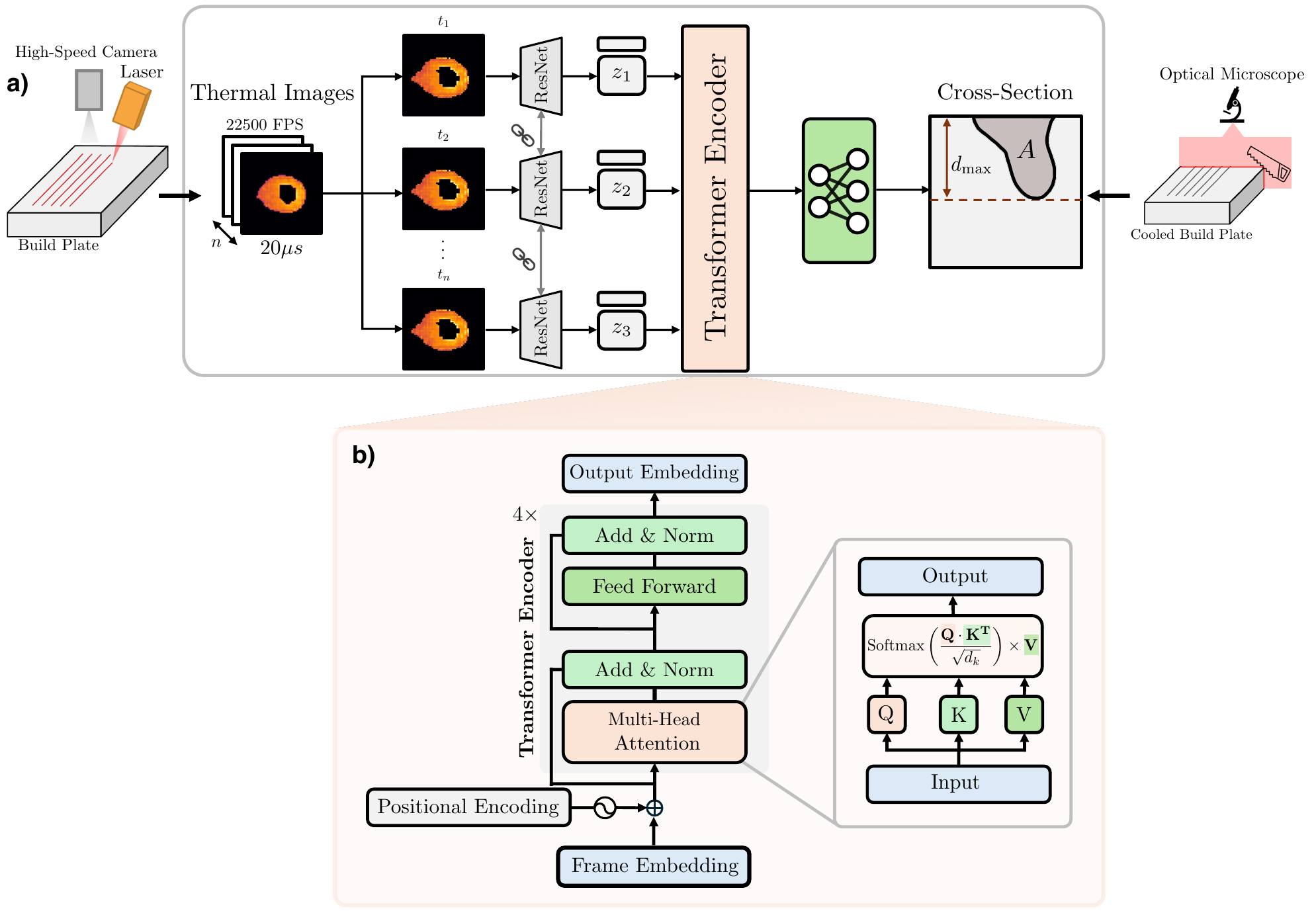}
\caption{\textbf{a)} A schematic of the architecture used for mapping the image sequence to the below surface melt pool morphology. In this schematic, $n$ observed melt pool thermal signatures are linked to the below surface melt pool shape. \textbf{b)} The architecture of the transformer encoder module. The core of the transformer encoder block is the multi-head attention step, which computes query ($Q$), key ($K$), and value ($V$) matrices that encode the relationships and dependencies between the frames of the video sequence.}
\label{fig:transformer_encoder}
\end{figure}

\subsection*{Self-Attention}
We develop a hybrid-CNN architecture to correlate thermal images of the melt pool surface to the morphology of the subsurface contour, while considering the  temporal variation of the melt pool. At the core of the transformer model is the concept of self-attention, which is reviewed briefly below \cite{vaswani2017attention, dosovitskiy2020image}.

The transformer encoder module implements attention to understand the temporal relationships across different frames of the input sequence. Attention is a deep learning paradigm for reasoning over sequential data while preserving long-range dependencies \cite{vaswani2017attention, bahdanau2014neural}. In recurrent models that do not apply attention, a hidden state vector is trained to store the relevant information contained within a given sequence \cite{chung2014empirical, graves2013generating}. However, at longer sequence lengths, the information available at early timesteps has a weaker influence on the model prediction than the more recent steps. Attention-based models learn a weighting of the relative importance of each element within the sequence, eliminating this effect. Within self-attention, the model learns what the most relevant aspects of the data are to each element in the sequence. 

This is implemented by developing a learned query, key and value matrix to compute the attention weights. The similarity between the query vector, $q$, and the key vector $k$, is used to weight the value vector.  

Specifically, the attention scores given a matrix of query vectors, $Q$, key vectors, $K$, and value vectors, $V$, is given as
\begin{equation}
\mathbf{a} = \mathrm{Softmax} \left ( \frac{\mathbf{Q} \cdot \mathbf{K^T}}{\sqrt{d_k}} \right ) \times V
\end{equation}

During multi-head self-attention, this approach is conducted in parallel over smaller sections of the intermediate query, where the query, key, and value vectors are divided along the channel dimension to create multiple instances of self-attention. In this scenario, the attention outputs are recombined from each head to form the final output.

\subsection*{Temporal Transformer Architecture}

While fully transformer-based architectures have been leveraged for computer vision tasks, the computational expense of performing spatial and temporal attention within a single transformer model is significant \cite{liu2022video, arnab2021vivit}. Additionally, transformer models sacrifice inductive biases for representational power, and therefore, perform optimally in large dataset scenarios. However, due to the smaller size of the annotated datasets that can be readily collected during process monitoring compared to traditional computer vision tasks \cite{deng2009imagenet}, we modify the transformer approach to balance the inductive biases provided through a CNN-based embedding with the representation power of the transformer paradigm. As the thermal image resolution is relatively small and the spatial relationships within the image remains consistent across the samples in the dataset, spatially applied attention is less valuable. However, the temporal behavior of the melt pool can be complex, and encoding this behavior via attention leverages the intended functionality of the transformer architecture to model long-range sequential relationships.  Therefore, the temporal relationships between these latent vectors are extracted with self-attention layers. In this implementation, each frame is first embedded with a ResNet architecture to a learned embedding space \cite{he2016deep}. Consistent with the terminology used for transformer architectures, the latent vector for each frame is referred to as a token \cite{vaswani2017attention}. Next, the token corresponding to each frame of the image sequence is concatenated with a sinusoidal positional embedding vector indicating where the frame lies within the sequence. A randomly initialized readout token is appended to the temporal sequence of tokens derived from the image sequence. This readout token is designed to store global information about the entire sequence. Finally, self-attention is computed across the sequence of tokens, and the embedded representation of the global behavior of the sequence is stored in this readout token. The readout token is then reconstructed to an estimate of the depth contour with a fully-connected network. This model is trained to minimize the mean squared error between the contour representation derived from experimental measurement, and the predicted contour representation.

\subsection*{Experimental Details} 
\label{subsec:experiment}
The 316L SS no-powder experiments were conducted on the TruPrint 3000 commercial L-PBF machine, which has a maximum laser power of 500~W and a maximum scanning velocity of 3~m/s. A high-speed color camera (Photron FASTCAM mini-AX200) with a high-magnification lens (Model K2 DistaMax™ with two NTX 2x extenders, standard objective lens, and Edmunds Optics tri-band filter 87-246) was mounted above the machine's front viewport, which contained two laser-blocking viewports (Edmunds Optics hot mirror 64-460). This imaging hardware is identical to the previous work by Myers and Quirarte \textit{et al.} \cite{myers2023high}. For all of the single beads, the high-speed camera was operated with a frame rate of 22,500~frames/sec, giving a field of view of 512~pixels by 512~pixels. The aperture was set near the half position, and the videos are taken with exposure times, $\Delta t$, of 4~$\mu$s and 20~$\mu$s. The hatch scans were imaged with a frame rate of 6,400~frames/sec to allow for a larger field of view (1024~pixels by 1024~pixels). In all cases, the instantaneous pixel field of view is approximately 5.6 $\mu$m on the build plate. Using the low exposure time (4~$\mu$s) allows for the measurement of peak temperatures in the melt pool, but due to the signal noise floor of the camera, lower temperatures are not measurable \cite{myers2023high}. Increasing the exposure time to 20~$\mu$s allows for the measurement of lower temperatures in the melt pool but results in saturation of the highest temperatures. The argon flow rate in the machine was set to 1.6~m/s, with an oxygen concentration between 0.2~\% and 0.4~\%. Each of the single beads were programmed to be 10~mm long with a lateral spacing of approximately 0.4~mm between them. The laser incidence angle is approximately 15 to 20 degrees, which causes the melt pool cross-section to be angled, especially in keyholing cases.

\begin{figure}[hbt!]

\includegraphics[width=1\linewidth]{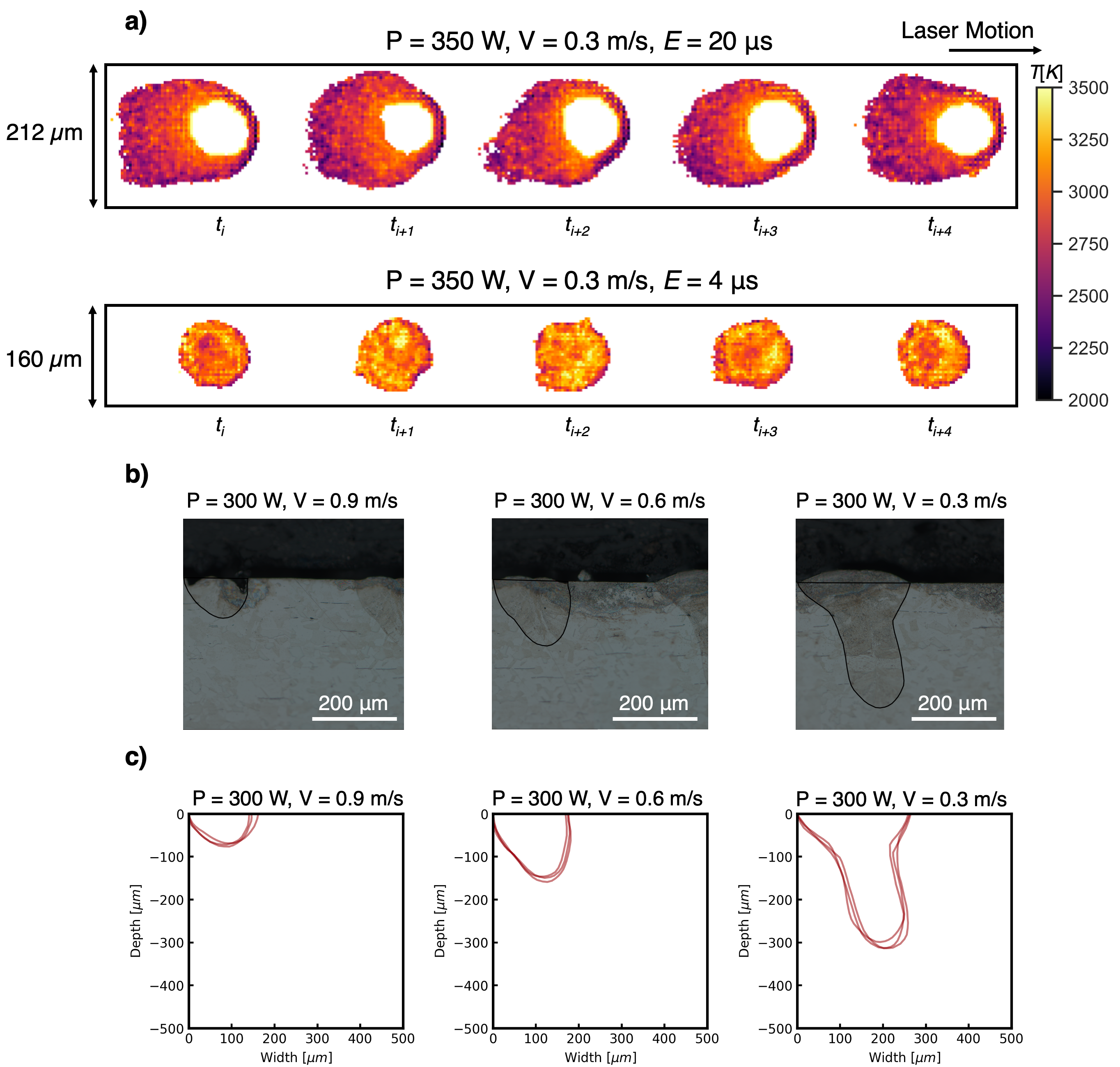}
\caption{\textbf{a)} The melt pool surface temperature images are sampled at intervals of 44 µs, corresponding to a frame rate of 22, 500 frames per second. The temperature distribution changes from frame to frame. At high exposure times, more of the melt pool structure is visible, but the temperature core of the melt pool near the laser is saturated. At lower exposure times, less of the melt pool is directly observable, but the center of the melt pool is no longer saturated. \textbf{b)} Optical micrographs of the cross-sectioned tracks at varying energy densities. \textbf{c)}The melt pool depth contour profiles for varying energy densities. There are slight variations due to the variability present between successive melt tracks. 
}
\label{fig:dataset_information}
\end{figure}

The 12-bit raw high-speed videos of the melt pool are then converted to both monochrome images and two-color thermal images. In both cases, the raw image is first demosaiced linearly, padded with zeroes, and cropped around the melt pool. The monochrome image is created by taking the red channel, multiplying it by a factor of 16, and writing it as a monochrome 16-bit TIF image. The ratiometric temperature is created by applying the two-color thermal imaging technique for a color camera using the red and green channels, which has been demonstrated for both LPBF and larger-scale directed energy deposition (DED) processes \cite{myers2023high,myers2023DED}. For the two-color thermal image, the temperature is only calculated at pixels with red and green 12-bit signals between 150 digital levels and 4000 digital levels.

After the single bead scans were run, the 316L SS plates were cross-sectioned, mounted, sanded, and polished. To identify the melt pool boundary, the polished samples were electro-etched following procedure 13b in the ASTM Standard Practice for Microetching Metals and Alloys (E-407). Then, the melt pool boundary profile was extracted from microscope images of the cross-sections by drawing around the edge of the melt pool in the ImageJ software. Four single beads were processed at each of the sixty-four combinations of power and velocity with eight powers ranging from 50~W to 400~W and eight velocities ranging from 0.3~m/s to 2.4~m/s. However, twenty-four melt pool cross-sections were not measurable due to either small melt pool size, over-etching, or under-etching.

\subsection*{Data Processing}
\label{subsec:dataprocessing}
Once the thermal images are extracted from the monitoring system, a series of image pre-processing steps are applied to prepare the dataset for machine learning inference. First, the thermal image is cropped to a 360$\times$360$\mu m$ area around the image to maintain spatial agreement between consecutive frames and remove extraneous blank pixels. Next, to reduce the temporal variation of the sequence attributable to noise, we apply a sequential moving window averaging operation. Specifically, each frame of the processed input video sequence is constructed as a spatial average of the $k$ raw frames immediately preceding and succeeding the current raw frame, $t_n$. Following the temporal averaging operation, overlapping sequences of each video are constructed as the model input. To create these overlapping sequences image frames $i$ to $i + m$, where $m$ is a tunable hyperparameter defining the window size of the sequence. This window is then shifted by one frame to form the next subsequence for model input, spanning frame $i+1$ to frame $i + m + 1$. With this data augmentation operation, we increase the amount of informative sequences that can be extracted from each combination of processing parameters. Samples from the thermal image dataset before processing is applied are shown in Figure \ref{fig:dataset_information}a).

Following the extraction of the cross-sectional melt pool contours, we featurize the contours for accurate prediction. Sample contours for the melt pool are shown in  Figure \ref{fig:dataset_information}b) for three laser conditions. By creating multiple tracks for each power-velocity combination, we observe variations in the morphology of the melt pool contour as the nominally specified parameters are held constant. In order to represent the contour, the melt pool cross-section is represented as a truncated signed distance function defined over a 2-D image \cite{curless1996volumetric}. Each pixel of this image stores the cartesian distance to the nearest point on the melt pool boundary. This distance is defined to be negative for points within the melt pool, and positive for points outside the melt pool. Therefore, the exact shape of the melt pool contour can be reconstructed by finding the iso-contour at 0 of the signed distance representation. In this work, we use a 64 $\times$ 64 image representation, where each pixel corresponds to a rectangular  6.25 $\mu m$ by 5.46 $\mu m$ area. This representation establishes the bounds of the predicted contour image as the maximum melt pool width and maximum melt pool depth observed over the entire dataset.

\section*{Results}
\label{sec:results}
\subsection*{Dataset Details}
\label{subsec:dataset}

To evaluate the performance of the proposed framework, we train the temporal transformer model to map video sequences of the melt pool thermal image to the corresponding melt pool depth cross-sections and examine the similarity of the reconstructed melt pools. The experimental monitoring dataset used for this work consists of 40 of the 64 unique combinations of laser power and laser scan speed used to print single bead tracks, with the remaining 24 omitted from the dataset due to the absence of measurable melt pool cross-sections.  This corresponds to 40 unique surface thermal image sequences at each exposure time, corresponding to 3425 image frames available at a 4 $\mu s$ exposure time and 3753 image frames available at a 20 $\mu s$ exposure time.

We separate the dataset into training and testing partitions based on a 75\% training, 25\% testing split. Specifically, 25\% of the power-velocity combinations present within the dataset are not considered during training to enable the evaluation of the performance of the model on unseen data. The train-test split is applied in the space of power-velocity combinations to avoid data leakage in the temporal sequences. 

A temporal averaging filter of $n$ = 5 frames is applied to the input video sequence to remove small-scale temporal fluctuations from the dataset. Following this, each video is subdivided into overlapping subsequences of 50 frames each to provide the model with temporal information about the melt pool motion.
The transformer model is trained for 50 epochs with a learning rate of $1\times 10^{-4}$ and a mini-batch size of 12. A weight decay factor of  $1\times 10^{-3}$ and a dropout factor of 0.1 is applied as a regularization factor to reduce overfitting. A sinusoidal positional encoding is applied within the model to provide information regarding the relative order of each frame within the input video sequence.  The temperature values within the surface thermal images are normalized by the maximum possible intensity of the 16-bit TIF representation to lie within the range [0, 1].

\subsection*{Metrics}

We evaluate model performance by examining the geometrical properties of the generated melt pool cross-section. Specifically, the melt pool area and melt pool depth is compared to a series of experimental measurements taken at similar processing conditions. We compare these two dimensional metrics for agreement, in addition to the intersection-over-union (IoU) score, which directly measures the overlap between the predicted and experimental ground truth melt pools within the dataset. The intersection over union score is defined as a ratio of the overlap area between the ground truth and predicted contours to the combined area of the ground truth and predicted contours. 

Additionally, we compute the mean absolute error (MAE) and the Pearson $R^2$ of the correlation between the ground truth and predicted dimensional metrics.

The Pearson $R^2$ coefficient is given by 
\begin{equation}
r = \frac{\sum (x_i - \bar{x})(y_i - \bar{y})}{\sqrt{\sum (x_i - \bar{x})^2}\sqrt{\sum (y_i - \bar{y})^2}}
\end{equation}
and the MAE is given by 
\begin{equation}
\mathrm{MAE} = \frac{1}{n} \sum_{i=1}^{n} |y_i - \hat{y}_i|
\end{equation}

Finally, we also establish metrics focused on preserving the shape of the melt pool contour. To do so, we calculate the MAE between the points defining the predicted melt pool contour, and the points defining the ground truth melt pool contour. To facilitate this, we resample each contour to consist of $n = 100$ horizontally equidistant points along the surface of the melt pool, where each horizontal co-ordinate component has a corresponding vertical co-ordinate component defining the depth of the melt pool at that point. The second metric used to benchmark the preservation of the melt pool contour shape is the Hausdorff distance \cite{huttenlocher1993comparing}. The Hausdorff distance is introduced to measure the maximum deviation between two contours, and is measured as the greatest distance from a point on one contour, to the nearest neighbor on the second contour.

The introduction of these metrics, as a complement to the IoU score, allow us to account for situations where the overall area of the melt pool may largely match the ground truth but localized deviations cause the shape of the melt pool to significantly differ in a localized region.

\subsection*{Baseline Performance}

\begin{figure}[hbt!]

\includegraphics[width=1\linewidth]{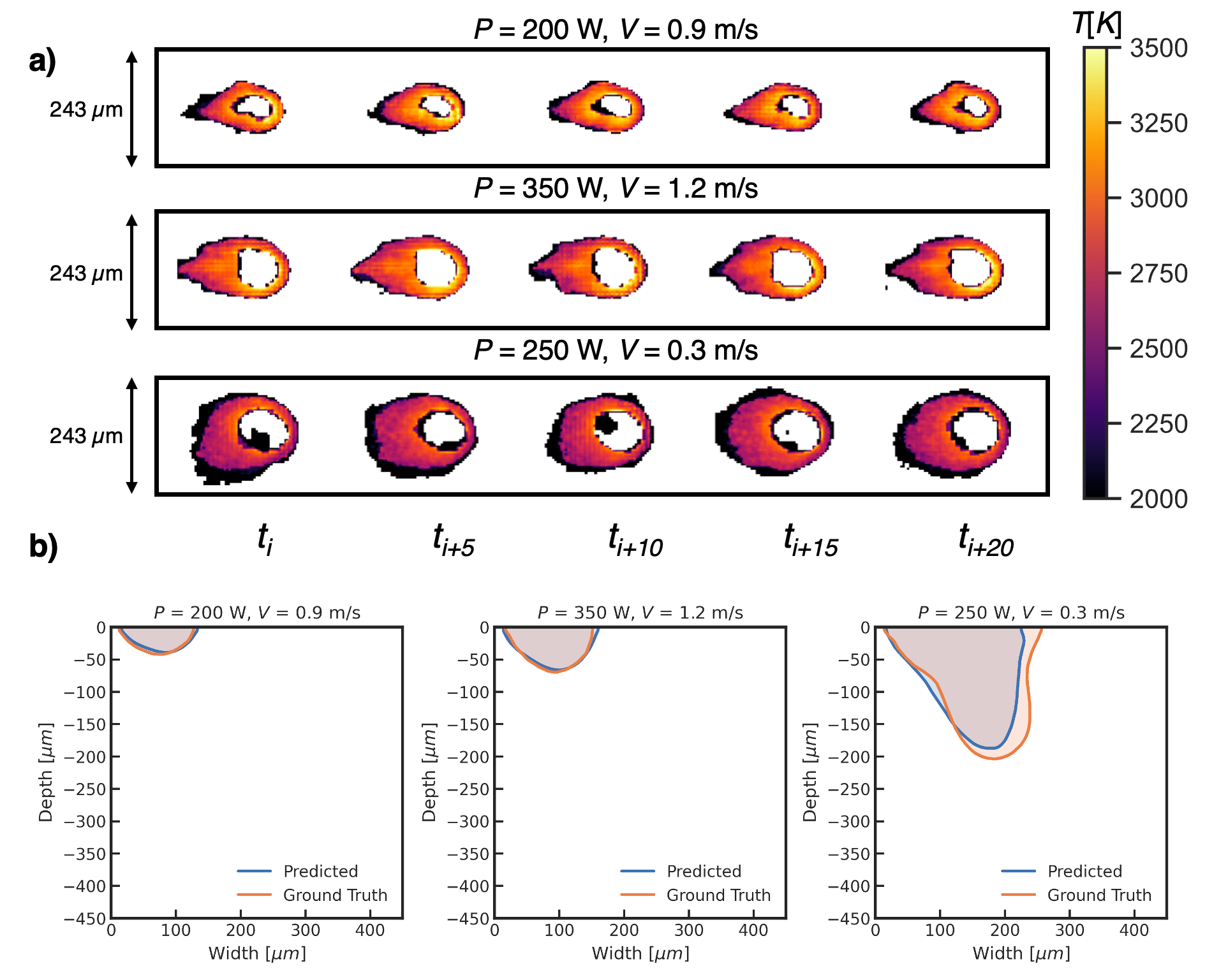}
\caption{\textbf{a)} Sample thermal images of the melt pool over time for P = 200 W, V = 0.9 m/s, P = 350 W, V = 1.2 m/s, and P = 250 W, V = 0.3 m/s  used as input for the depth contour prediction model. A moving average window is applied temporally for a period of 220 $\mu s$ ($n$ = 5), and a sequence length of 50 frames is used for the prediction. A 20 $\mu s$ exposure time is used as input to the depth contour prediction model. \textbf{b) }A comparison of the depth contours predicted by the temporal transformer model to the ground truth depth contours.  
}
\label{fig:qualitative_results}
\end{figure}

First, we evaluate the performance of our proposed model on the varying monitoring configurations used to collect the surface thermal image dataset. The corresponding IoU scores are computed for four different configurations of the melt pool image, contrasting the performance with monochrome images to the performance with temperature images, and contrasting the performance of images collected with a 4 $\mu s$ exposure time to those collected with a 20 $\mu s$ exposure time. We examine these results for one sample configuration in Figure \ref{fig:qualitative_results}, where temperature estimates at a 20 $\mu s$ exposure time are provided for analysis. Figure \ref{fig:qualitative_results}a) demonstrates the temporal variation of the melt pool once a moving average window of $n$ = 5 frames is applied to the image sequence. The core structure of the melt pool is preserved over time, and variation is localized to the lower temperature fringes of the melt pool surface image.  Figure \ref{fig:qualitative_results}b) demonstrates sample prediction contours for the three sample melt pool sequences shown in Figure \ref{fig:qualitative_results}a). Close agreement is observed qualitatively between the morphology of the ground truth melt pool cross-section and the predicted melt pool cross-section. 

\begin{figure}[hbt!]

\centering
\includegraphics[width=1\linewidth]{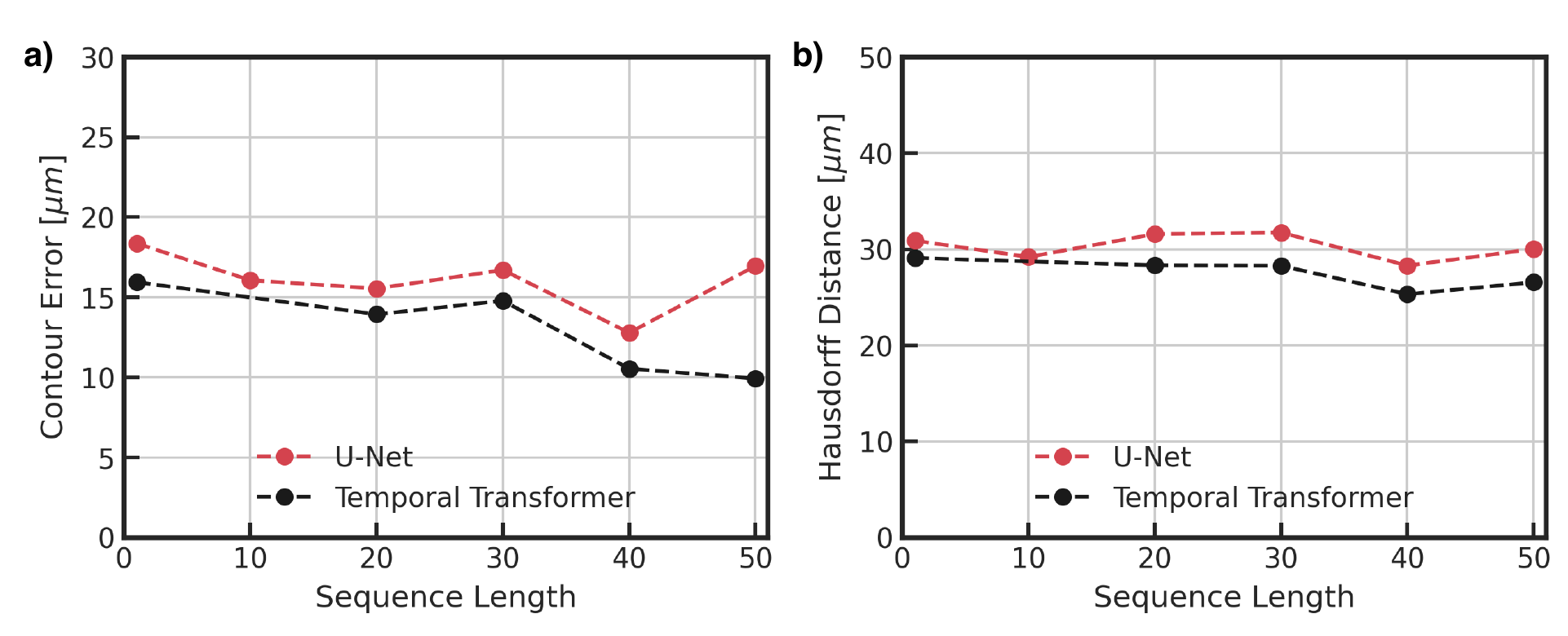}

\caption{ A comparison of the U-Net and Temporal Transformer prediction performance as a function of sequence length. \textbf{a)} The mean absolute error of the resampled predicted contour points, compared to the resampled ground truth contour points as a function of the sequence length. \textbf{(b)} The Hausdorff distance between the predicted contour points, and the ground truth contour points.  }
\label{fig:seq_length_comparison}
\end{figure}

The performance of the transformer model on each configuration is listed in Table \ref{table:monochromevtemperature}, where the best-performing configuration is indicated in bold. Notably, we observe that the most accurate predictions are obtained with the temperature estimates derived from the ratiometric temperature images. The IoU of the predicted cross-sections appears is almost identical between an exposure time of $E = 4 \mu s$, and an exposure time of $E = 20 \mu s$. With this configuration, the trained model achieves the lowest error between the predicted and experimentally observed melt pool dimensions, and the largest IoU score.

\begin{table}[htbp]
\centering
\small
\caption{Contour Evaluation Results}
\label{table:contour_evaluation}
\vspace{0.5mm}
\resizebox{\textwidth}{!}{\begin{tabular}{lcccccc}\toprule
& \multicolumn{3}{c}{$E = 4 \; \mu s$} & \multicolumn{3}{c}{$E = 20 \; \mu s$} \\
\cmidrule(lr){2-4} \cmidrule(lr){5-7}
         Model Architecture & IoU $\uparrow$ & Hausdorff Distance $\downarrow$ & Contour Error $\downarrow$ & IoU $\uparrow$ & Hausdorff Distance $\downarrow$ & Contour Error $\downarrow$  \\\midrule
Temporal Transformer & 0.76 $\pm$ 0.04  & \textbf{20.09 $\pm$ 3.96}  & \textbf{9.92 $\pm$ 3.00}  & 0.77 $\pm$ 0.01  &\textbf{ 19.23 $\pm$ 1.30}  & \textbf{8.14 $\pm$ 1.80 } \\ 
ViT & \textbf{0.77 $\pm$ 0.02}  & 25.55 $\pm$ 0.35  & 11.23 $\pm$ 0.32  & \textbf{0.79 $\pm$ 0.01}  & 24.56 $\pm$ 0.02  & 9.17 $\pm$ 0.74  \\ 
UNet & 0.76 $\pm$ 0.02  & 26.54 $\pm$ 3.01  & 11.76 $\pm$ 0.72  & 0.76 $\pm$ 0.06  & 30.47 $\pm$ 2.12  & 13.37 $\pm$ 1.79  \\ 
\bottomrule
\end{tabular}}
\end{table}

\begin{table}[htbp]
\caption{Thermal Image Processing Results}
\label{table:monochromevtemperature}
\centering
\begin{tabular}{ccc}\toprule
Exposure Time, $E$ & \multicolumn{1}{c}{$4 \; \mu s$} & \multicolumn{1}{c}{$ 20 \; \mu s$} \\
\cmidrule(lr) {1-1} \cmidrule(lr){2-2} \cmidrule(lr){3-3}
         Image Format & IoU $\uparrow$ & IoU $\uparrow$ \\\midrule
Monochrome & 0.73 $\pm$ 0.03  & 0.75 $\pm$ 0.01  \\ 
Temperature & \textbf{0.77 $\pm$ 0.04}  & \textbf{ 0.77 $\pm$ 0.01}  \\

\bottomrule

\end{tabular}
\end{table}

Similarly, we examine the influence of the sequence length chosen for the sub-sequence definition on the prediction quality of the melt pool cross-sections (Figure \ref{fig:seq_length_comparison}). As the sequence length increases, the performance of the model improves, increasing from an area correlation $R^2$ of 0.78 at a sequence length of a single frame to an area correlation $R^2$ of 0.88 at a sequence length of 40 frames. From this analysis, at longer sequence lengths the model is able to observe longer-range temporal dependencies, and is less sensitive to the instantaneous fluctuations of the melt pool temperature due to the short time-scale phenomenon occurring in the process. While a longer sequence length may lead to increased performance, it also limits the timescale at which laser control would be possible with this system.

\subsection*{Model Comparison}

To evaluate the performance of the proposed model, we benchmark the observed performance against architectures designed to only consider the spatial component of the melt pool structure, neglecting the temporal information. Specifically, we compare the performance of a U-Net architecture and a Vision Transformer architecture to the performance observed with our proposed model \cite{dosovitskiy2020image, ronneberger2015u}.  These models have shown demonstrated success in related computer vision tasks similar to the task presented here, such as image segmentation and depth estimation. To adapt the task for these 2-D architectures, we apply the same data augmentation strategy to divide each image sequence into multiple sub-sequences of a specified length. However, the images within each sub-sequence are then mean-aggregated in time to construct a single 2-D input image for each sub-sequence. 

\begin{figure}[hbt!]

\includegraphics[width=1\linewidth]{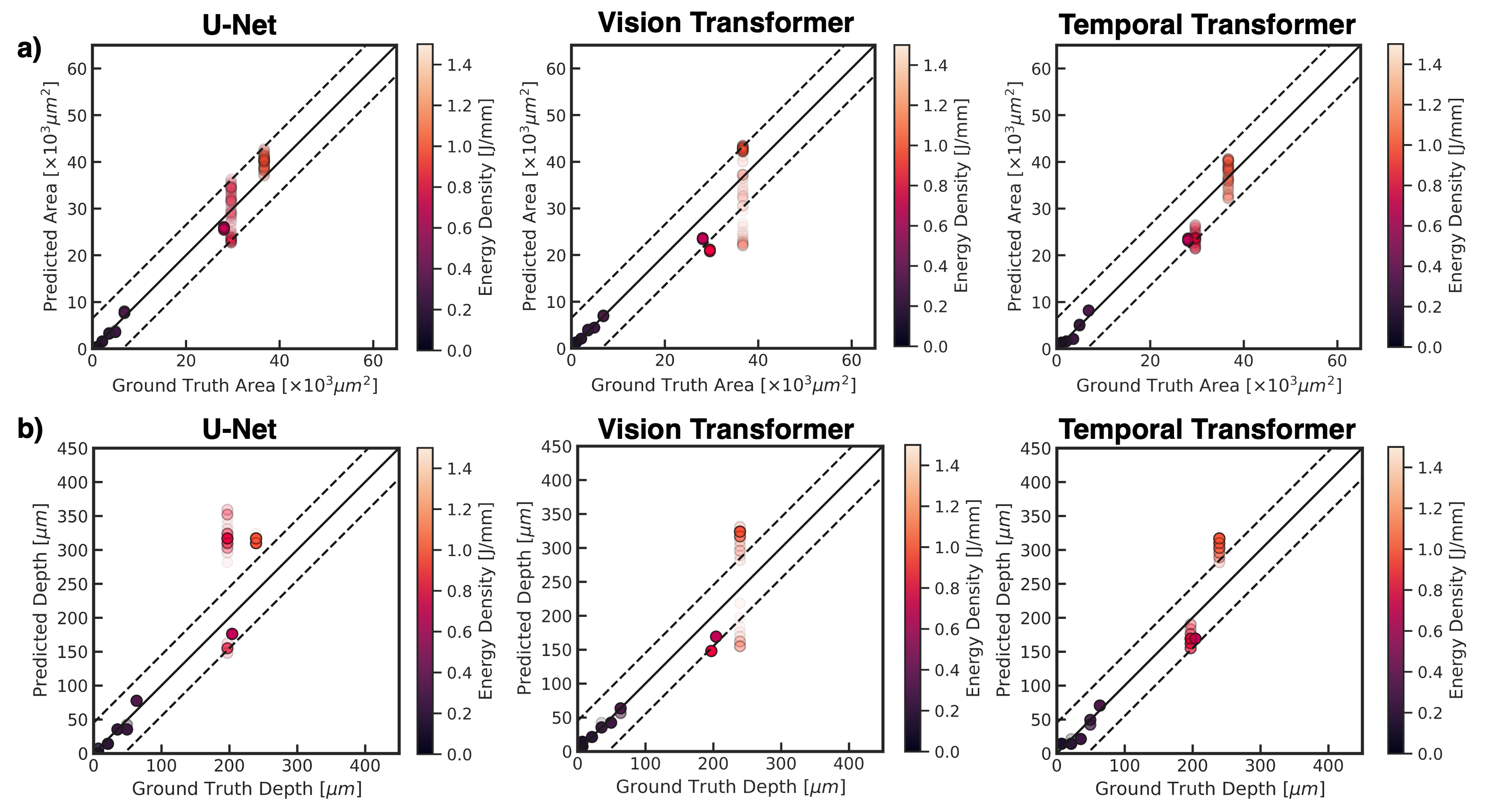}
\caption{A comparison of the extracted melt pool dimensions for contour predictions on the test partition of the experimental dataset, for three different model architectures. a) The area correlation between the ground truth and predicted melt pool contours, for the U-Net, Vision Transformer, and Temporal Transformer models. The dotted lines indicate a deviation of $\pm 6500 \; \mu m^{2}$ from the ideal prediction. b) The depth correlation between the ground truth and predicted melt pool contours, for the U-Net, Vision Transformer, and Temporal Transformer models. The dotted lines indicate a deviation of $\pm 45 \; \mu m$ from the ideal prediction.}
\label{fig:model_comparison}
\end{figure}

Each model is trained for 50 epochs on the same train-test dataset split, and the pre-processing steps are held constant. More details regarding the specific configuration of each model are provided in the Appendix. We evaluate the performance of each of the three architectures for two different exposure times on the surface temperature estimates (Table \ref{table:contour_evaluation}), where the best-performing configuration is also indicated in bold. 

We first examine the comparison between a U-Net architecture for this task, and the implemented transformer architecture. We find that our model out-performs the U-Net on the contour shape metrics, and achieves comparable accuracy on the IoU metric. Specifically, the $\pm$$1\sigma$  confidence interval of the temporal transformer network falls directly within the confidence interval of the U-Net architecture's performance on the IoU metric. We graphically compare the performance of the three models by comparing the experimental melt pool dimensions with the dimensions of the corresponding predicted melt pool (Figure \ref{fig:model_comparison}). When examining the model predictions on a process parameter basis, we observe that at higher energy densities, the variation within the predicted melt pool dimensions becomes more significant (Figure \ref{fig:model_comparison}). However, the implemented transformer architecture has much less variation in the predictions produced over time by a single model. Notably, our model is also significantly smaller with 7.5 $\times 10^6$ parameters compared to the 8 $\times 10^7$ parameters within the implemented standard U-Net model.  

We also examine the influence of applying attention temporally by evaluating the performance of our model against a conventional Vision Transformer (ViT) that only operates spatially on individual 2-D images. This Vision Transformer model applies attention in the spatial axis, dividing the input image into sub-windows, or patches, of size 16 x 16, and learning how each patch attends to the other patches within the image \cite{dosovitskiy2020image}. This paradigm allows for prediction over complex images with features at multiple scales, however, at the cost of computational requirements that grow quadratically with the number of patches. Therefore, we compare the choice of applying attention spatially as opposed to temporally by benchmarking the performance of our model against a conventional vision transformer. Following this comparison, we note that our implemented model also outperforms the ViT model on the melt pool shape metrics, and achieves equivalent performance on the Intersection-over-Union benchmark. Similarly to the U-Net benchmark discussed earlier, we are able to achieve this performance with fewer parameters by combining the inductive biases within convolutional networks with the sequence learning abilities of transformer models.

\begin{figure}[hbt!]

\includegraphics[width=1\linewidth]{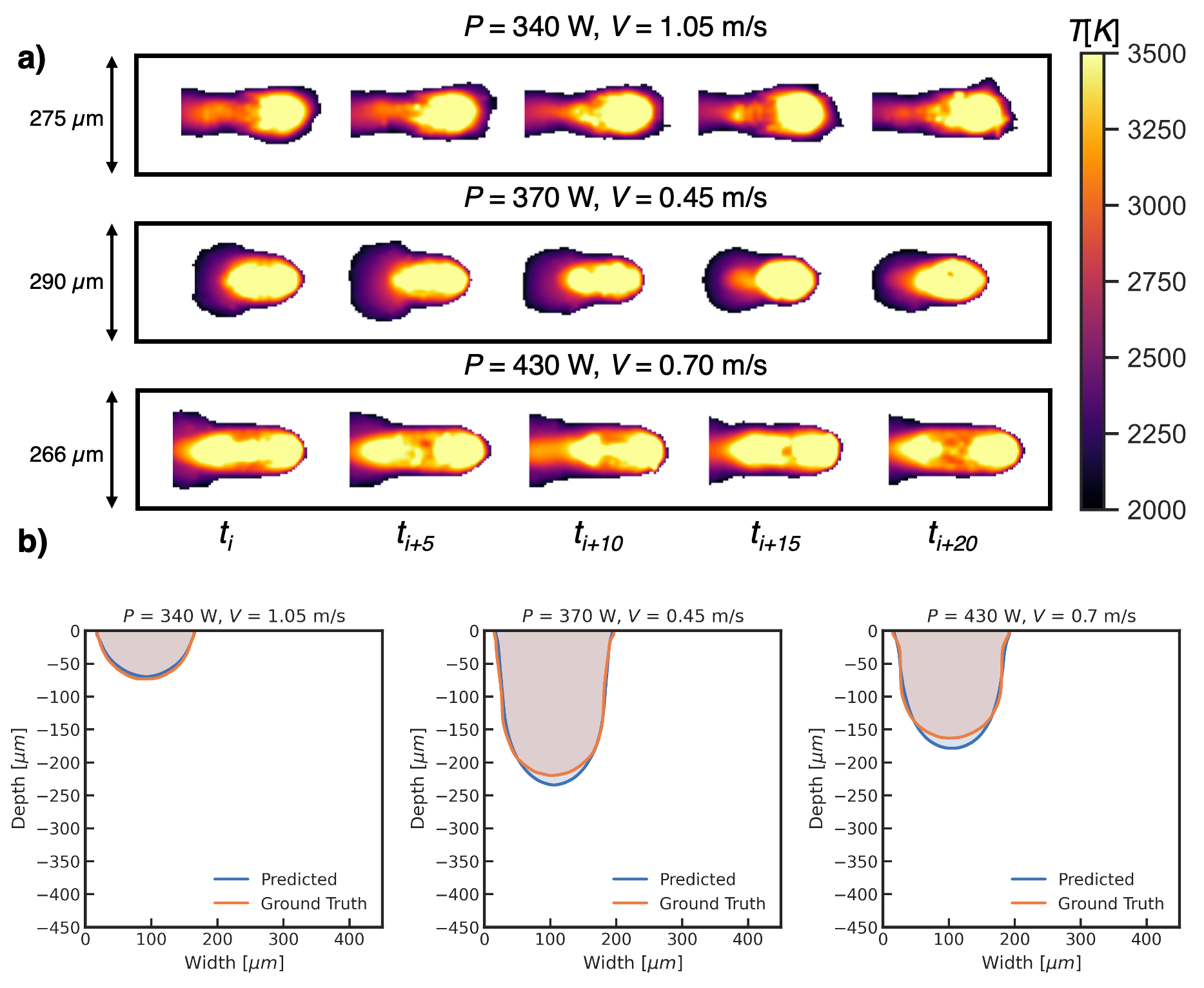}
\caption{\textbf{a)} Sample thermal images of the FLOW-3D simulated melt pool over time for P = 340 W, V = 1.05 m/s, P = 370 W, V = 0.45 m/s, and P = 430 W, V = 0.7 m/s  used as input for the depth contour prediction model. \textbf{b) }A comparison of the depth contours predicted by the temporal transformer model to the depth contours obtained via FLOW-3D simulation.  
}
\label{fig:flow3d_qualitative_results}
\end{figure}

\label{subsec:metrics}
\subsection*{Pre-Training}

\label{subsec:pretraining}
To alleviate the data requirements for training the model, we investigate the integration of simulation data with experimental observations. The process of cross-sectioning and etching samples becomes time-consuming at large scales, therefore, reducing the number of annotated labels required will increase the efficiency of this framework for future applications.



The approach implemented for integrating simulation data with experimental data is based upon transfer learning, where models are first trained via simulation results and subsequently fine-tuned on experimental observations. In this framework, a model is first trained to predict the melt pool cross-sectional morphology using simulation-based surface images and cross-sections. This study is performed for both data produced from high-fidelity, multiphysics simulations of the melt pool behavior with the CFD software FLOW-3D\textsuperscript{®} \cite{FLOW-3D}, and data produced from simplified analytical models of the heat conduction process \cite{eagar1983temperature}.  This enables us to evaluate the need for performing more computationally expensive simulations to create data, as opposed to using less accurate low-fidelity models. To evaluate the performance of the high-fidelity simulation data, we first create a dataset of 300  SS316L single-track bare plate runs using FLOW-3D.  FLOW-3D is a multiphysics package which solves the coupled partial differential equations governing mass transfer, heat transfer, and momentum transfer within the melt pool. FLOW-3D incorporates considerations for the physical phenomena taking place during the melting process, including phase change, laser reflection, and evaporation. A more detailed description of these considerations is provided within \cite{myers2023high, ogoke2023inexpensive}. 

These simulations are constructed with FLOW-3D (Version 12.0, Release 7) to cover the power-velocity space with a laser power range of 130 W to a laser power of 500 W at an increment of 30 W, and a velocity range of 0.3 m/s to 1.5 m/s at an increment of 0.05 m/s. The material parameters of the simulation were validated following the process described in \cite{myers2023high}.

To extract the equivalent melt pool contour from the simulation data, we define a time-aggregated composite melt pool that compiles the maximum temperature seen for a specific grid cell over the entire length of the simulation. Next, we apply an iso-contour at a threshold $T^*$, which is defined to be halfway between the solidus and liquidus temperatures. For each simulation, we extract the surface temperature profiles by projecting the three-dimensional free surface of the melt pool onto a flat plane.

$$T_{\mathrm{surface}} = \max_{\mathbf{z}} \; T(\mathbf{x},\mathbf{y},\mathbf{z})$$

To ensure consistency with the experimental monitoring data, the input thermal images and target depth contours are rescaled to match the experimental monitoring resolution. Specifically, a $320 \mu m \times 320 \mu m$ area around the laser is extracted and upscaled by a factor of two to obtain a 64$\times$64 pixel image resolution. Further interpolation and grid rescaling is applied to ensure the spatial resolution of the simulated surface thermal images matches the 5.6 $\mu m$ pixel resolution of the experimental thermal images. 

We train the model for 100 epochs on the FLOW-3D dataset to examine its performance in simulation, free of measurement noise. The results of the training process are shown in Figure \ref{fig:flow3d_qualitative_results}. Notably, we achieve an $R^2$ of 0.98 on the melt pool depth and width computed from the predicted contours, and an IoU of 0.99 for the overlap between the predicted and simulation contours.

\begin{figure}[hbt!]

\centering
\includegraphics[width=1\linewidth]{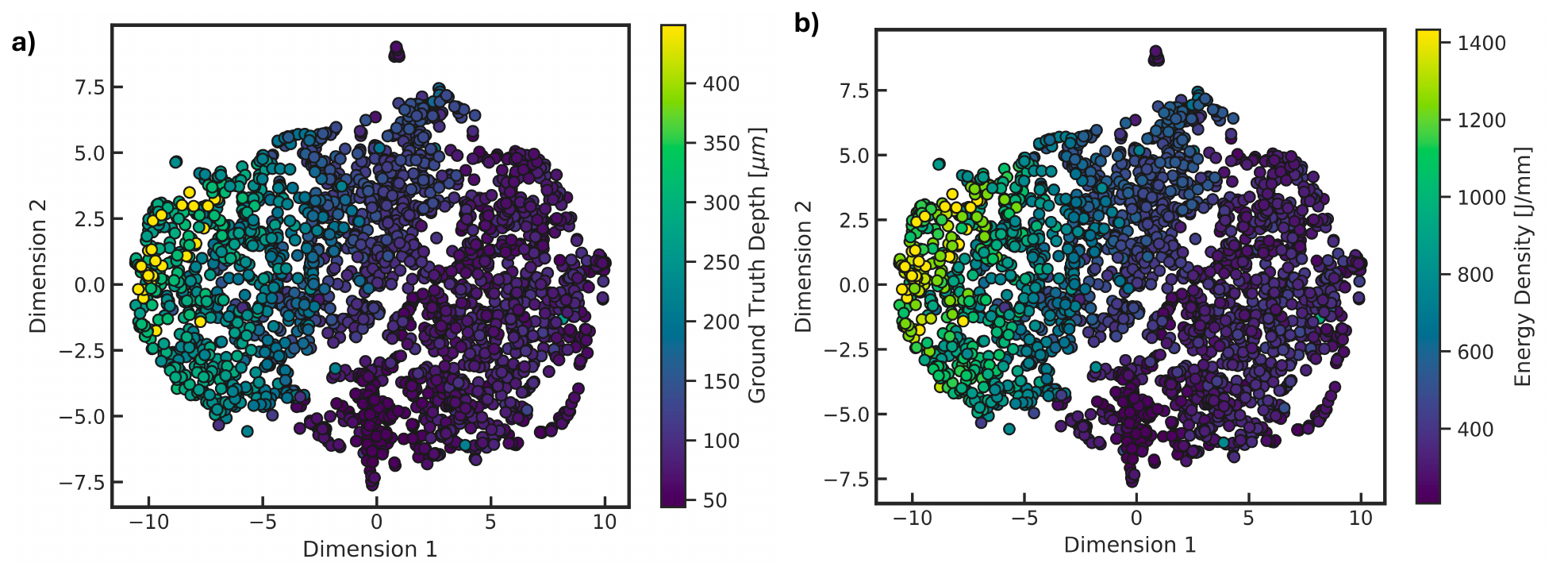}

\caption{\textbf{a)} A \textit{t}-SNE embedding plot based on the latent space embedding learned within the transformer directly before the prediction head, on the dataset of FLOW-3D simulations. A trend is observed where the samples with the largest depth are concentrated at the top right of the plot. \textbf{b)} A similar trend is observed with increasing energy density.\textbf{ c)} The \textit{t}-SNE embedding reveals that the model has the largest error at low energy density and depth values.}
\label{fig:tsne}
\end{figure}

To examine the learned embedding space of the model trained on simulated surface temperatures, we perform dimensionality reduction on the 128-dimensional readout token optimized during the training process. To examine how this token changes throughout the dataset in a visually intuitive manner, we apply the dimensionality reduction algorithm \textit{t}-distributed Stochastic Neighbor Embedding (\textit{t}-SNE), to produce a reduced order 2-dimensional vector representation for each sample in the dataset \cite{van2008visualizing}.  The correlation of these low-dimensional representations with the properties of the melt pool are shown in Figure \ref{fig:tsne}, where each point represents the compressed read-out token associated with a specific data sample. In  Figure \ref{fig:tsne}a), intuitive trends are observed within the 2-dimensional latent space. Specifically, samples with similar depths in the unseen test dataset are observed to be clustered together in the latent space, and a smooth transition between the shallow melt pools and deeper melt pools is also apparent. In Figure \ref{fig:tsne}b), these trends are also shown to hold for the energy density of each melt pool, also demonstrating that the model is able to learn generalizable and intuitive representations of the underlying physical mechanisms of melt pool formation.

\subsubsection*{Model Transfer to Experimental Data}
\begin{figure}[hbt!]

\includegraphics[width=1\linewidth]{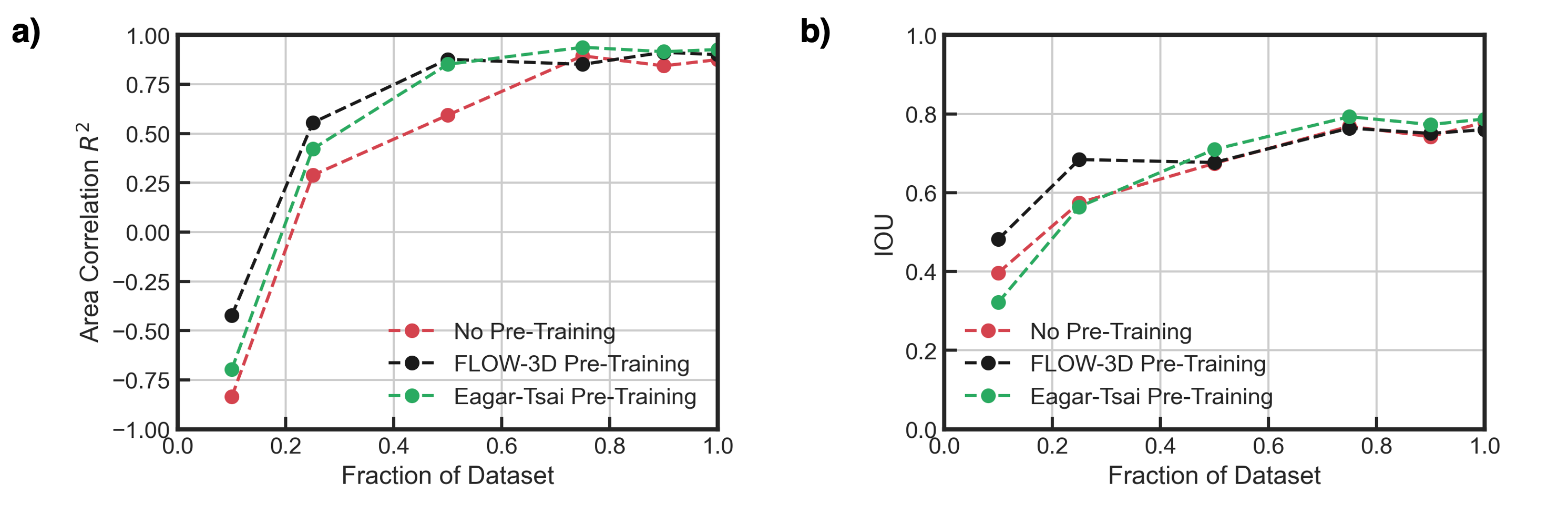}
\caption{\textbf{a)} A comparison of the Intersection-over-Union (IoU) metric as a function of the amount of data used for model training. These comparisons are made for both a model trained with randomly initialized weights, a model that has been transfer learned from FLOW-3D simulated melt pools, and a model that has been transfer learned from Eagar-Tsai simulated melt pools. \textbf{b)} A comparison of the correlation between the area extracted from the predicted cross-section to the area extracted from the ground truth cross-section, a function of the amount of data used for model training. These comparisons are made for both a model trained with randomly initialized weights, a a model that has been transfer learned from FLOW-3D simulated melt pools, and a model that has been transfer learned from Eagar-Tsai simulated melt pools.}
\label{fig:pretraining_comparison}
\end{figure}

We also create a dataset of low-fidelity simulations using the Eagar-Tsai model to simulate the melt pool. The Eagar-Tsai model has more restrictive assumptions than the high-fidelity simulation models. Specifically, the Eagar-Tsai model considers heat conduction as the only mode of heat dissipation, assumes that the thermal properties of the material remain constant with temperature, and neglects phase change phenomena \cite{eagar1983temperature}. Due to these assumptions, the melt pool temperatures computed are up to 10$\times$ higher than those reported in experiment \cite{myers2023high}. However, the Eagar-Tsai model has a semi-analytical solution which is less expensive to compute than a FLOW-3D simulation at the same mesh resolution \cite{wolfer2019fast, ogoke2023inexpensive}. To assess the performance of pre-training on this melt pool model for transfer learning, we create a dataset of 270 SS316L simulations and compute the surface temperature and depth contour with the same procedure described earlier for the FLOW-3D dataset. The semi-analytical equation for the Eagar-Tsai model is reproduced in Equation \ref{eq:ET}, where $A$ is the laser absorptivity, $\rho$ is the density of the material, $c_p$ is the heat capacity, $k$ is the thermal conductivity, $D$ is the thermal diffusivity defined as $D = \frac{k}{\rho c_p}$, $\sigma$ is the radius of the laser, $V$ is the velocity of the laser, and $P$ is the power of the laser. The parameters used to compute the solution at different power-velocity combinations are reported in \ref{sec:ET Details}.

\begin{equation}
\label{eq:ET}
T(x, y, z) = \frac{A P}{\rho c_{p} \sqrt{4 \pi^3 D}} \times \int_{0}^{\Delta t^{(i)}} \frac{\bar{\tau}^{-1/2}}{ \sigma^2 + 2 D \bar{\tau} }  \exp \left\{ -\frac{(x + V \bar{\tau})^2 + y^2 }{2 \sigma^2 + 4 D \bar{\tau}} - \frac{z^2}{4 D \bar{\tau}} \right \} d \bar{\tau}
\end{equation}

To apply the information learned by the model in simulation to the experimentally collected dataset, we first trained the model for 100 epochs on the simulation dataset as described earlier. Next, we fine-tune the trained model on a subset of the training partition of the single-track melt pool cross-sections for 10 epochs. We initialize the model with the weights learned from the simulation data training process, and continue the training process with a randomly selected sample of the training partition of the experimental dataset. Next, we evaluate the performance of the fine-tuned model on the original test partition of the dataset as the fraction of training data available as input varied. In Figure \ref{fig:pretraining_comparison}a), the correlation between the ground truth and predicted melt pool cross-sectional areas is shown to increase dramatically when a model pre-trained on simulation data is used for prediction, compared to a network with randomly initialized weights. We observe a greater increase in accuracy in the low dataset regime for the FLOW-3D pre-trained model when compared to the Eagar-Tsai pretrained model. However, as more data becomes available for the fine-tuning process, similar increases in accuracy are seen for both the FLOW-3D and Eagar-Tsai pre-trained models. A similar effect is observed in Figure \ref{fig:pretraining_comparison}b) for the IoU score of the predicted cross-sections. Therefore, fine-tuning a FLOW-3D pre-trained model on only 10 unique process parameter combinations enables predictions with an area correlation $R^2$ of 0.56, and increasing the amount of data to 20 unique process-parameter cross-sections enables prediction with an area correlation $R^2$ of 0.85.

\label{subsec:featurization}

\subsubsection*{Hatch Spacing Analysis}

\begin{figure}[hbt!]

\includegraphics[width=1\linewidth]{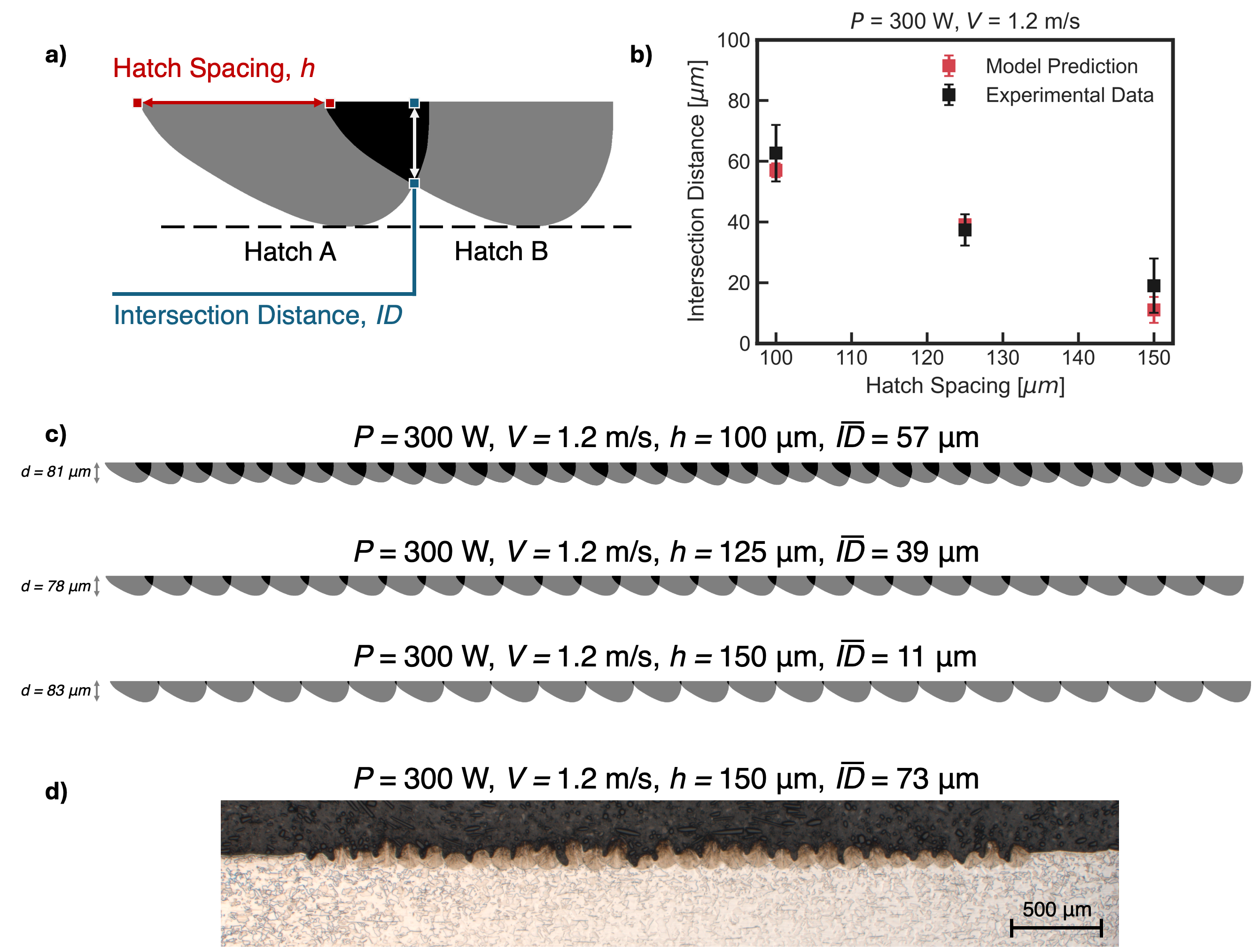}
\caption{We investigate the ability of the machine learning model to predict information about the melt pool overlap over different hatch spacing values. \textbf{a) }The calculation of the intersection distance from the melt pool overlap observed at a certain hatch spacing.\textbf{ b)} A comparison of the intersection distance observed after applying the transformer model to predict the melt pool morphology.\textbf{ c)} Qualitative examples of the melt pool overlap behavior predicted at different hatch spacing values and energy densities. d) A sample optical micrograph of the overlapping scan tracks at $h$ = 125~$\mu m$, $P$ = 300~W, $V$ = 1.2~m/s.}
\label{fig:hatch_spacing}
\end{figure}

Finally, we investigate the application of our model towards predicting the overlap between melt pool contours observed during multi-track printing. The failure of the melt tracks to correctly overlap can create lack-of-fusion porosity. Therefore, we seek to understand whether the predicted melt pool contours can be used as an \textit{in-situ} monitoring tool for identifying overlap between successive hatches. To do so, we collect high-speed two-color thermal images of the melt pool following the specifications described above, at a frame rate of 6400 FPS. Similarly, we again cross-section, polish, and etch the solidified build plate. With this dataset, we leverage our model trained on single-track runs at varying laser energy densities to predict the melt pool depth observed at each hatch from the corresponding thermal image sequence. The results of this analysis are shown in Figure \ref{fig:hatch_spacing}. We use the distance from the surface of the melt pool to the intersection of the hatches as a metric to investigate if the overlap is correctly captured. In Figure \ref{fig:hatch_spacing}b), we show that the statistics of the experimentally observed intersection distances match directly with the intersection distances observed in experiment. In Figure \ref{fig:hatch_spacing}c) and Figure \ref{fig:hatch_spacing}d), qualitative examples of the predicted and experimentally observed cross-sectioned tracks are shown. The close agreement between the ML predicted melt pool overlap and the overlap demonstrated in experiment demonstrates promise towards using this methodology as a tool for \textit{in-situ} defect detection.

\section*{Conclusion}

In  this work, a method for predicting the melt pool cross-sectional morphology from \textit{in-situ} high-speed two-color thermal imaging of the melt pool surface was created. This is accomplished through the implementation of a hybrid CNN-Transformer model. By  decomposing the model architecture into the spatial and temporal components of the prediction task, we leverage the advantages of both convolutional and attention-based architectures. The stricter inductive bias of CNNs is used to extract spatial information from individual melt pool frames using a ResNet backbone. Next, a transformer model learned the temporal relationships between these encoded thermal image frames to discover temporal dependencies. Comparisons are made between the performance of architectures that do not consider temporal information, with our proposed model demonstrating greater stability, as shown in Figure \ref{fig:model_comparison}.

The performance of this framework is evaluated based on the dimensions of the predicted melt pool, as well as shape-dependent metrics such as the Intersection-over-Union score. The implemented model is able to predict the contour within an average deviation of 10 $\mu m$, and a maximum deviation of 20 $\mu m$. By comparing the effect of the optical monitoring configuration, we note that the use of ratiometric temperature estimates increases the IoU over monochromatic images from 0.73 to 0.77. Finally, we evaluate the utility of transfer learning to reduce the amount of experimental data required for robust model training. By applying this strategy, predictions within an $R^2$ of 0.87 are achieved with a 50 \% reduction in the number of annotated cross-sections necessary for analysis. The development of this model enables the prediction of the melt pool cross-section directly from \textit{in-situ} monitoring information, allowing L-PBF practitioners to have robust early detection of lack-of-fusion and porosity estimates. Future extensions of this work could examine the temporal information that is learned by a sequential model at higher sampling rates for greater insight into the physics underlying the temporal oscillation of the melt pool. Additionally, experiments with powder applied to the build plate can yield additional insight into the variability of the observed melt pool during the build process.

\label{sec:conclusion}
\section*{Acknowledgements}
The authors acknowledge use of the Materials Characterization Facility at Carnegie Mellon University supported by grant MCF-677785. Research was sponsored by the Army Research Laboratory, USA and was accomplished under Cooperative Agreement Number W911NF-20-2-0175. The views and conclusions contained in this document are those of the authors and should not be interpreted as representing the official policies, either expressed or implied, of the Army Research Laboratory or the U.S. Government. The U.S. Government is authorized to reproduce and distribute reprints for Government purposes notwithstanding any copyright notation herein.

This work is also supported by Army ERDC Grant No. 1990749. A.J.M., G.Q., and a portion of this research is supported by the National Science Foundation Graduate Research Fellowship Program under Grant No. DGE2140739 \& DGE1745016. Any opinions, findings, and conclusions or recommendations expressed in this material are those of the author(s) and do not necessarily reflect the views of the National Science Foundation. We thank Scott Kram for his help with the L-PBF machine. We also acknowledge the use of the NextManufacturing facilities at Carnegie Mellon University.

\appendix

 \bibliographystyle{elsarticle-num} 
 \bibliography{cas-refs}

\begin{thebibliography}{10}
\expandafter\ifx\csname url\endcsname\relax
  \def\url#1{\texttt{#1}}\fi
\expandafter\ifx\csname urlprefix\endcsname\relax\def\urlprefix{URL }\fi
\expandafter\ifx\csname href\endcsname\relax
  \def\href#1#2{#2} \def\path#1{#1}\fi

\bibitem{li2020review}
Y.~Li, Z.~Feng, L.~Hao, L.~Huang, C.~Xin, Y.~Wang, E.~Bilotti, K.~Essa, H.~Zhang, Z.~Li, et~al., A review on functionally graded materials and structures via additive manufacturing: from multi-scale design to versatile functional properties, Advanced Materials Technologies 5~(6) (2020) 1900981.

\bibitem{reeves2011additive}
P.~Reeves, C.~Tuck, R.~Hague, Additive manufacturing for mass customization, in: Mass customization, Springer, 2011, pp. 275--289.

\bibitem{cunningham2017analyzing}
R.~Cunningham, A.~Nicolas, J.~Madsen, E.~Fodran, E.~Anagnostou, M.~D. Sangid, A.~D. Rollett, Analyzing the effects of powder and post-processing on porosity and properties of electron beam melted ti-6al-4v, Materials Research Letters 5~(7) (2017) 516--525.

\bibitem{mower2016mechanical}
T.~M. Mower, M.~J. Long, Mechanical behavior of additive manufactured, powder-bed laser-fused materials, Materials Science and Engineering: A 651 (2016) 198--213.

\bibitem{spierings2013fatigue}
A.~B. Spierings, T.~L. Starr, K.~Wegener, Fatigue performance of additive manufactured metallic parts, Rapid prototyping journal (2013).

\bibitem{lewandowski2016metal}
J.~J. Lewandowski, M.~Seifi, Metal additive manufacturing: a review of mechanical properties, Annual review of materials research 46 (2016).

\bibitem{tofail2018additive}
S.~A. Tofail, E.~P. Koumoulos, A.~Bandyopadhyay, S.~Bose, L.~O’Donoghue, C.~Charitidis, Additive manufacturing: scientific and technological challenges, market uptake and opportunities, Materials today 21~(1) (2018) 22--37.

\bibitem{yadroitsev2015hierarchical}
I.~Yadroitsev, P.~Krakhmalev, I.~Yadroitsava, Hierarchical design principles of selective laser melting for high quality metallic objects, Additive Manufacturing 7 (2015) 45--56.

\bibitem{zhang2018evolution}
T.~Zhang, H.~Li, S.~Liu, S.~Shen, H.~Xie, W.~Shi, G.~Zhang, B.~Shen, L.~Chen, B.~Xiao, et~al., Evolution of molten pool during selective laser melting of ti--6al--4v, Journal of Physics D: Applied Physics 52~(5) (2018) 055302.

\bibitem{khairallah2016laser}
S.~A. Khairallah, A.~T. Anderson, A.~Rubenchik, W.~E. King, Laser powder-bed fusion additive manufacturing: Physics of complex melt flow and formation mechanisms of pores, spatter, and denudation zones, Acta Materialia 108 (2016) 36--45.

\bibitem{matthews2020controlling}
M.~Matthews, T.~Roehling, S.~Khairallah, T.~Tumkur, G.~Guss, R.~Shi, J.~Roehling, W.~Smith, B.~Vrancken, R.~Ganeriwala, et~al., Controlling melt pool shape, microstructure and residual stress in additively manufactured metals using modified laser beam profiles, Procedia Cirp 94 (2020) 200--204.

\bibitem{li2018residual}
C.~Li, Z.~Liu, X.~Fang, Y.~Guo, Residual stress in metal additive manufacturing, Procedia Cirp 71 (2018) 348--353.

\bibitem{koepf20183d}
J.~A. Koepf, M.~R. Gotterbarm, M.~Markl, C.~K{\"o}rner, 3d multi-layer grain structure simulation of powder bed fusion additive manufacturing, Acta Materialia 152 (2018) 119--126.

\bibitem{gong2014analysis}
H.~Gong, K.~Rafi, H.~Gu, T.~Starr, B.~Stucker, Analysis of defect generation in ti--6al--4v parts made using powder bed fusion additive manufacturing processes, Additive Manufacturing 1 (2014) 87--98.

\bibitem{ning2019analytical}
J.~Ning, W.~Wang, B.~Zamorano, S.~Y. Liang, Analytical modeling of lack-of-fusion porosity in metal additive manufacturing, Applied Physics A 125~(11) (2019) 1--11.

\bibitem{ronneberg2020revealing}
T.~Ronneberg, C.~M. Davies, P.~A. Hooper, Revealing relationships between porosity, microstructure and mechanical properties of laser powder bed fusion 316l stainless steel through heat treatment, Materials \& Design 189 (2020) 108481.

\bibitem{rice1997limitations}
R.~Rice, Limitations of pore-stress concentrations on the mechanical properties of porous materials, Journal of Materials Science 32~(17) (1997) 4731--4736.

\bibitem{wilson2021combined}
A.~E. Wilson-Heid, A.~M. Beese, Combined effects of porosity and stress state on the failure behavior of laser powder bed fusion stainless steel 316l, Additive Manufacturing 39 (2021) 101862.

\bibitem{bayat2019keyhole}
M.~Bayat, A.~Thanki, S.~Mohanty, A.~Witvrouw, S.~Yang, J.~Thorborg, N.~S. Tiedje, J.~H. Hattel, Keyhole-induced porosities in laser-based powder bed fusion (l-pbf) of ti6al4v: High-fidelity modelling and experimental validation, Additive Manufacturing 30 (2019) 100835.

\bibitem{tang2017prediction}
M.~Tang, P.~C. Pistorius, J.~L. Beuth, Prediction of lack-of-fusion porosity for powder bed fusion, Additive Manufacturing 14 (2017) 39--48.

\bibitem{rosenthal1941mathematical}
D.~Rosenthal, Mathematical theory of heat distribution during welding and cutting, aw s, Jourual, May (1941).

\bibitem{eagar1983temperature}
T.~Eagar, N.~Tsai, et~al., Temperature fields produced by traveling distributed heat sources, Welding journal 62~(12) (1983) 346--355.

\bibitem{cheng2019computational}
B.~Cheng, L.~Loeber, H.~Willeck, U.~Hartel, C.~Tuffile, Computational investigation of melt pool process dynamics and pore formation in laser powder bed fusion, Journal of Materials Engineering and Performance 28 (2019) 6565--6578.

\bibitem{markl2016multiscale}
M.~Markl, C.~K{\"o}rner, Multiscale modeling of powder bed--based additive manufacturing, Annual Review of Materials Research 46 (2016) 93--123.

\bibitem{gaikwad2022multi}
A.~Gaikwad, R.~J. Williams, H.~de~Winton, B.~D. Bevans, Z.~Smoqi, P.~Rao, P.~A. Hooper, Multi phenomena melt pool sensor data fusion for enhanced process monitoring of laser powder bed fusion additive manufacturing, Materials \& Design 221 (2022) 110919.

\bibitem{Tian2020Deep}
Q.~Tian, Q.~Tian, S.~Guo, E.~Melder, L.~Bian, W.~Guo, Deep learning-based data fusion method for in situ porosity detection in laser-based additive manufacturing, Journal of Manufacturing Science and Engineering (2020).
\newblock \href {https://doi.org/10.1115/1.4048957} {\path{doi:10.1115/1.4048957}}.

\bibitem{taherkhani2022unsupervised}
K.~Taherkhani, C.~Eischer, E.~Toyserkani, An unsupervised machine learning algorithm for in-situ defect-detection in laser powder-bed fusion, Journal of Manufacturing Processes 81 (2022) 476--489.

\bibitem{strayer2022accelerating}
S.~T. Strayer, W.~J.~F. Templeton, F.~X. Dugast, S.~P. Narra, A.~C. To, Accelerating high-fidelity thermal process simulation of laser powder bed fusion via the computational fluid dynamics imposed finite element method (cifem), Additive Manufacturing Letters 3 (2022) 100081.

\bibitem{hemmasian2023surrogate}
A.~Hemmasian, F.~Ogoke, P.~Akbari, J.~Malen, J.~Beuth, A.~B. Farimani, Surrogate modeling of melt pool temperature field using deep learning, Additive Manufacturing Letters 5 (2023) 100123.

\bibitem{ogoke2023inexpensive}
F.~Ogoke, Q.~Liu, O.~Ajenifujah, A.~Myers, G.~Quirarte, J.~Beuth, J.~Malen, A.~B. Farimani, Inexpensive high fidelity melt pool models in additive manufacturing using generative deep diffusion, arXiv preprint arXiv:2311.16168 (2023).

\bibitem{jadhav2023stressd}
Y.~Jadhav, J.~Berthel, C.~Hu, R.~Panat, J.~Beuth, A.~B. Farimani, Stressd: 2d stress estimation using denoising diffusion model, Computer Methods in Applied Mechanics and Engineering 416 (2023) 116343.

\bibitem{ogoke2021thermal}
F.~Ogoke, A.~B. Farimani, Thermal control of laser powder bed fusion using deep reinforcement learning, Additive Manufacturing 46 (2021) 102033.

\bibitem{chen2022data}
F.~Chen, M.~Yang, W.~Yan, Data-driven prognostic model for temperature field in additive manufacturing based on the high-fidelity thermal-fluid flow simulation, Computer Methods in Applied Mechanics and Engineering 392 (2022) 114652.

\bibitem{buchbinder2011high}
D.~Buchbinder, H.~Schleifenbaum, S.~Heidrich, W.~Meiners, J.~B{\"u}ltmann, High power selective laser melting (hp slm) of aluminum parts, Physics Procedia 12 (2011) 271--278.

\bibitem{gobert2018application}
C.~Gobert, E.~W. Reutzel, J.~Petrich, A.~R. Nassar, S.~Phoha, Application of supervised machine learning for defect detection during metallic powder bed fusion additive manufacturing using high resolution imaging., Additive Manufacturing 21 (2018) 517--528.

\bibitem{scime2018anomaly}
L.~Scime, J.~Beuth, Anomaly detection and classification in a laser powder bed additive manufacturing process using a trained computer vision algorithm, Additive Manufacturing 19 (2018) 114--126.

\bibitem{shevchik2018acoustic}
S.~A. Shevchik, C.~Kenel, C.~Leinenbach, K.~Wasmer, Acoustic emission for in situ quality monitoring in additive manufacturing using spectral convolutional neural networks, Additive Manufacturing 21 (2018) 598--604.

\bibitem{scime2019melt}
L.~Scime, J.~Beuth, Melt pool geometry and morphology variability for the inconel 718 alloy in a laser powder bed fusion additive manufacturing process, Additive Manufacturing 29 (2019) 100830.

\bibitem{larsen2022deep}
S.~Larsen, P.~A. Hooper, Deep semi-supervised learning of dynamics for anomaly detection in laser powder bed fusion, Journal of Intelligent Manufacturing 33~(2) (2022) 457--471.

\bibitem{vaswani2017attention}
A.~Vaswani, N.~Shazeer, N.~Parmar, J.~Uszkoreit, L.~Jones, A.~N. Gomez, {\L}.~Kaiser, I.~Polosukhin, Attention is all you need, Advances in neural information processing systems 30 (2017).

\bibitem{dosovitskiy2020image}
A.~Dosovitskiy, L.~Beyer, A.~Kolesnikov, D.~Weissenborn, X.~Zhai, T.~Unterthiner, M.~Dehghani, M.~Minderer, G.~Heigold, S.~Gelly, et~al., An image is worth 16x16 words: Transformers for image recognition at scale, arXiv preprint arXiv:2010.11929 (2020).

\bibitem{bahdanau2014neural}
D.~Bahdanau, K.~Cho, Y.~Bengio, Neural machine translation by jointly learning to align and translate, arXiv preprint arXiv:1409.0473 (2014).

\bibitem{chung2014empirical}
J.~Chung, C.~Gulcehre, K.~Cho, Y.~Bengio, Empirical evaluation of gated recurrent neural networks on sequence modeling, arXiv preprint arXiv:1412.3555 (2014).

\bibitem{graves2013generating}
A.~Graves, Generating sequences with recurrent neural networks, arXiv preprint arXiv:1308.0850 (2013).

\bibitem{liu2022video}
Z.~Liu, J.~Ning, Y.~Cao, Y.~Wei, Z.~Zhang, S.~Lin, H.~Hu, Video swin transformer, in: Proceedings of the IEEE/CVF conference on computer vision and pattern recognition, 2022, pp. 3202--3211.

\bibitem{arnab2021vivit}
A.~Arnab, M.~Dehghani, G.~Heigold, C.~Sun, M.~Lu{\v{c}}i{\'c}, C.~Schmid, Vivit: A video vision transformer, in: Proceedings of the IEEE/CVF international conference on computer vision, 2021, pp. 6836--6846.

\bibitem{deng2009imagenet}
J.~Deng, W.~Dong, R.~Socher, L.-J. Li, K.~Li, L.~Fei-Fei, Imagenet: A large-scale hierarchical image database, in: 2009 IEEE conference on computer vision and pattern recognition, Ieee, 2009, pp. 248--255.

\bibitem{he2016deep}
K.~He, X.~Zhang, S.~Ren, J.~Sun, Deep residual learning for image recognition, in: Proceedings of the IEEE conference on computer vision and pattern recognition, 2016, pp. 770--778.

\bibitem{myers2023high}
A.~J. Myers, G.~Quirarte, F.~Ogoke, B.~M. Lane, S.~Z. Uddin, A.~B. Farimani, J.~L. Beuth, J.~A. Malen, High-resolution melt pool thermal imaging for metals additive manufacturing using the two-color method with a color camera, Additive Manufacturing (2023) 103663.

\bibitem{myers2023DED}
A.~J. Myers, G.~Quirarte, J.~L. Beuth, J.~A. Malen, Two-color thermal imaging of the melt pool in powder-blown laser-directed energy deposition, Additive Manufacturing 78 (2023) 103855.

\bibitem{curless1996volumetric}
B.~Curless, M.~Levoy, A volumetric method for building complex models from range images, in: Proceedings of the 23rd annual conference on Computer graphics and interactive techniques, 1996, pp. 303--312.

\bibitem{huttenlocher1993comparing}
D.~P. Huttenlocher, G.~A. Klanderman, W.~J. Rucklidge, Comparing images using the hausdorff distance, IEEE Transactions on pattern analysis and machine intelligence 15~(9) (1993) 850--863.

\bibitem{ronneberger2015u}
O.~Ronneberger, P.~Fischer, T.~Brox, U-net: Convolutional networks for biomedical image segmentation, in: Medical image computing and computer-assisted intervention--MICCAI 2015: 18th international conference, Munich, Germany, October 5-9, 2015, proceedings, part III 18, Springer, 2015, pp. 234--241.

\bibitem{FLOW-3D}
I.~Flow~Science, \href{https://www.flow3d.com/}{FLOW-3D, Version~12.0}, Santa Fe, NM (2019).
\newline\urlprefix\url{https://www.flow3d.com/}

\bibitem{van2008visualizing}
L.~Van~der Maaten, G.~Hinton, Visualizing data using t-sne., Journal of machine learning research 9~(11) (2008).

\bibitem{wolfer2019fast}
A.~J. Wolfer, J.~Aires, K.~Wheeler, J.-P. Delplanque, A.~Rubenchik, A.~Anderson, S.~Khairallah, Fast solution strategy for transient heat conduction for arbitrary scan paths in additive manufacturing, Additive Manufacturing 30 (2019) 100898.

\end{thebibliography}
\newpage





\label{sec:sample:appendix}
\section{Experimental Data}

\begin{figure}[hbt!]

\includegraphics[width=1\linewidth]{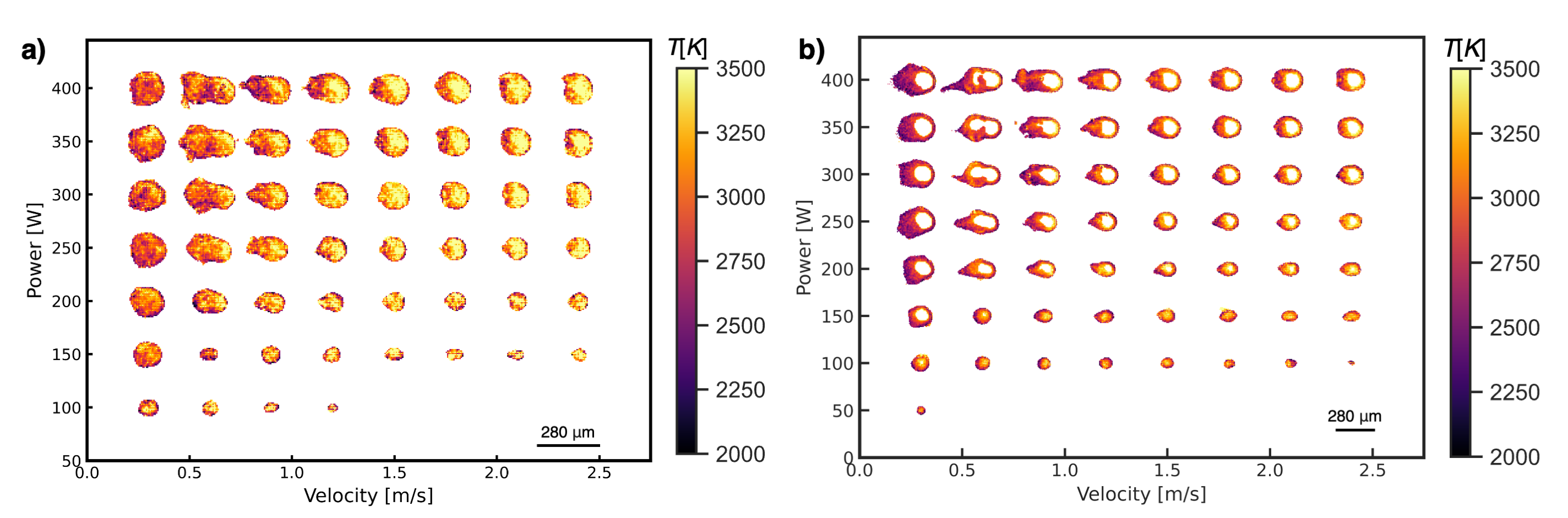}
\caption{Sample thermal images after ratiometric conversion for the power-velocity combinations studied in this work. \textbf{a)} Thermal images at an exposure time of E = 4 $\mu s$. \textbf{b)} Thermal images at an exposure time of E = 20 $\mu s$.}
\label{fig:thermal_image}
\end{figure}

\begin{figure}[hbt!]

\includegraphics[width=1\linewidth]{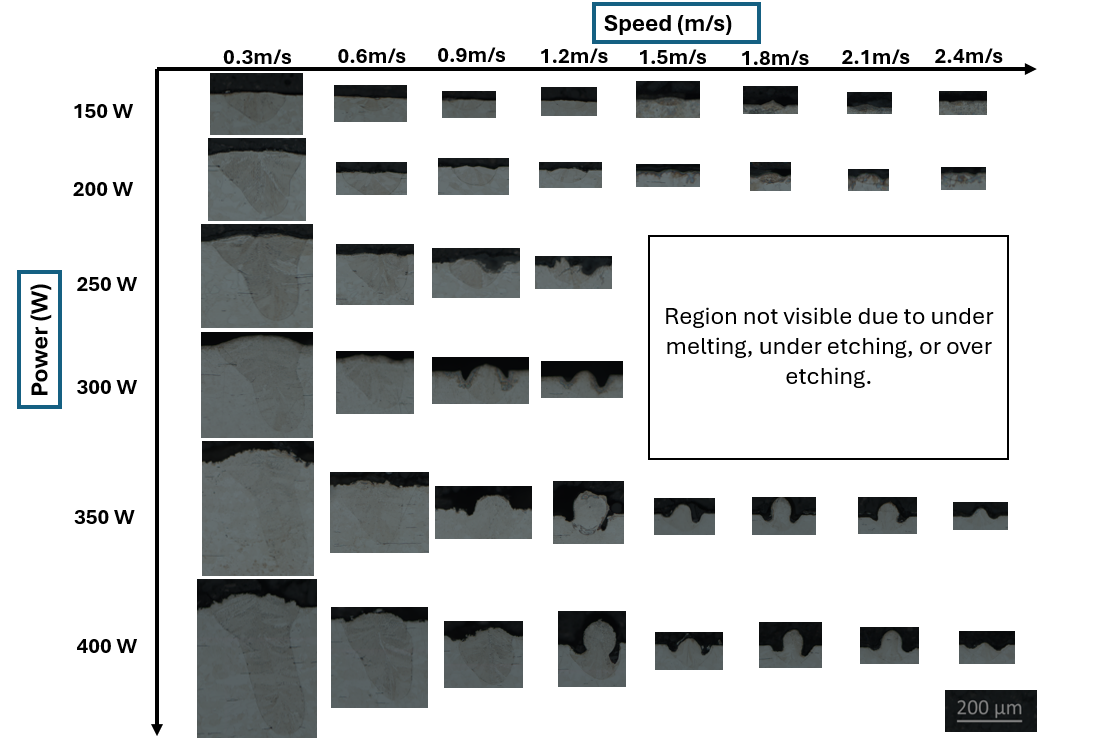}
\caption{Optical micrographs of the cross-sectional tracks at varying energy densities, ranging from Power = 150 - 400W and Velocity = 0.3 - 2.4 m/s.}
\label{fig:micrograph_figures}
\end{figure}

\section{FLOW-3D Simulation Details}
We simulate a single-track bare plate melting simulation for each unique power-velocity combination in the FLOW-3D dataset. The simulated build plate is $1000 \mu m \times 600 \mu m \times 400 \mu m$, in the $x$, $y$, and $z$ directions respectively. The $x$-axis indicates the laser direction of travel, while the $y$-axis lies along the width of the generated melt pool, and the $z$-axis is defined as the vertical axis. Each simulation is carried out for 500 $\mu s$ and data is reported at 5 $\mu s$ intervals, for a total of 100 time-dependent snapshots for each power-velocity combination. The power-velocity combinations used are identical to those reported in \cite{ogoke2023inexpensive} for the SS316L dataset. The material parameters used for simulation are reproduced in Table \ref{table:flow3dss316lmaterialparam}.

\begin{table}[htbp!]
\centering
\caption{Material Parameters used to simulate the SS316L melt pool with the FLOW-3D model.}
\begin{tabular}{@{}lllll@{}}
\toprule
Parameter                        & Value                                            & Units                   &  &  \\ \midrule
Density, $\rho$, 298 K              & 7950                                             & kg/m$^3$ &  &  \\
Density, $\rho$, 1923 K             & 6765                                            & kg/m$^3$ &  &  \\
Specific Heat, $C_v$, 298 K         & 470                                              & J/kg/K                  &  &  \\
Specific Heat, $C_v$, 1923 K        & 1873                                              & J/kg/K                  &  &  \\
Vapor Specific Heat, $C_{v, vapor}$ & 449                                              & J/kg/K                  &  &  \\
Thermal Conductivity, $k$, 298 K    & 13.4                                                & W/m/K                   &  &  \\
Thermal Conductivity, $k$, 1923 K  & 30.5                                            & W/m/K                   &  &  \\
Viscosity, $\eta$                        & 0.008                                         & kg/m/s                  &  &  \\
Surface Tension, $\sigma$          & 1.882 &          $kg/s^2$          &  &  \\
Liquidus Temperature, $T_L$       & 1723                                             & K                       &  &  \\
Solidus Temperature, $T_S$        & 1658                                             & K                       &  &  \\

Fresnel Coefficient, $\epsilon$         &0.15  &         -           &  &  \\

Accommodation Coefficient, $a$          &0.25  & -                    &  &  \\

Latent Heat of Fusion, $\Delta H_f$            & 2.6 $\times 10^5$ & J/kg                    &  &  \\
Latent Heat of Vaporization, $\Delta H_v$    & 7.45 $\times 10^6$ & J/kg                    &  &  \\ \bottomrule
\end{tabular}
\label{table:flow3dss316lmaterialparam}
\end{table}

\section{Eagar-Tsai Simulation Details}
\label{sec:ET Details}


We also simulate a single-track bare plate melting simulation for each unique power-velocity combination in the Eagar-Tsai dataset. The simulated build plate is $1000 \mu m \times 600 \mu m \times 400 \mu m$, in the $x$, $y$, and $z$ directions respectively. The simulation is solved to simulate the laser moving for 200 $\mu s$ at the specified velocity. The power of the laser is defined to vary from 50 W to 450 W at increments of 400 W. The velocity of the laser is defined to vary from 0.1 m/s to 1.55 m/s in increments of 0.05 m/s. The material parameters used for simulation are reproduced in Table \ref{table:ss316lmaterialparam}.

\begin{table}[htbp!]
\centering

\caption{Material Parameters used to simulate the SS316L melt pool with the Eagar-Tsai model.}
\begin{tabular}{@{}llll@{}}
\toprule
Parameter                        & Value                                            & Units                   &   \\ \midrule

Density, $\rho$             & 8030                                            & kg/m$^3$ &    \\

Thermal Conductivity, $k$  &   16.2                                     & W/m/K                   &    \\

Melting Point, $T_m$        & 1658                                             & K                       &   \\
Beam Radius, $\sigma$        &      50                                        & $\mu m$                     &    \\
Absorptivity, $A$         &0.3  &         -           &    \\

Heat Capacity, $c_p$         &500  &         -           J/kg~K &   
 \\ \bottomrule
\end{tabular}
\label{table:ss316lmaterialparam}
\end{table}

\FloatBarrier

\section{Experimental Data Variability}

\begin{figure}[hbt!]

\centering
\includegraphics[width=1\linewidth]{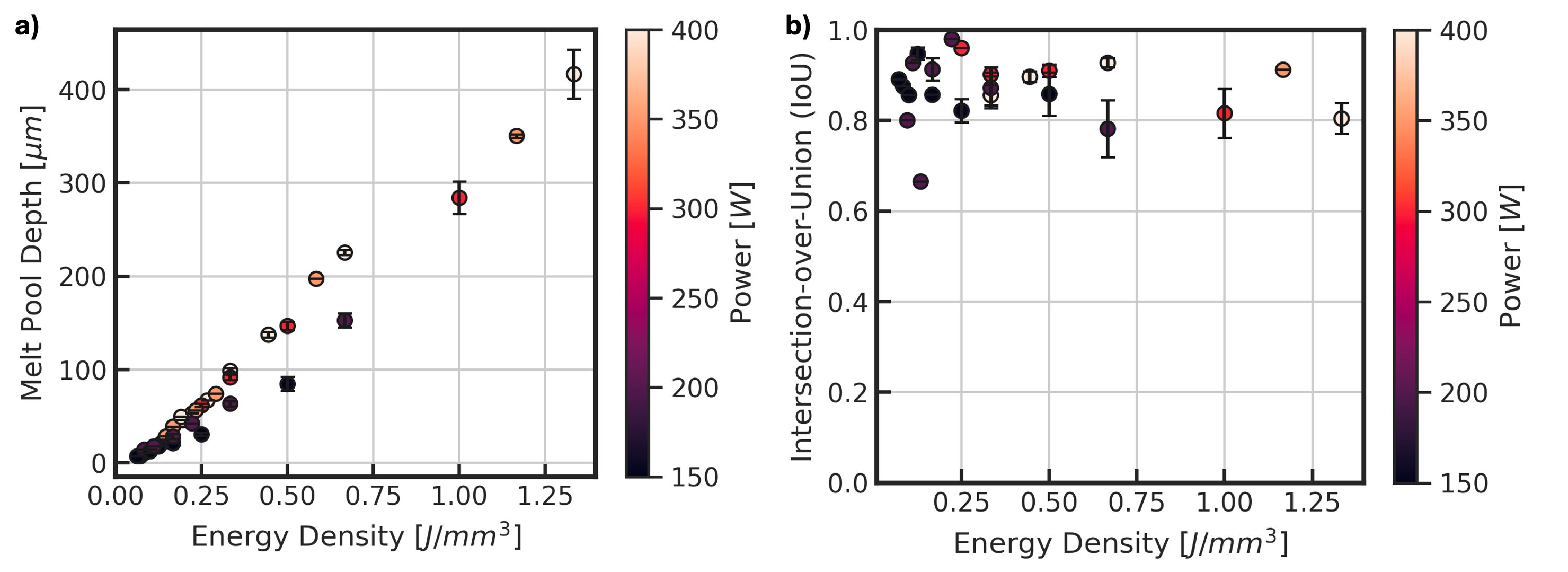}

\caption{The variation in the peak cross-section depth \textbf{(a)} and the cross-section melted shape \textbf{(b)} as a function of energy density. The melted depth variation is measured by taking the mean and standard deviation of the peak melt pool depth extracted at a specific power and velocity combination. The shape variation is measured by first calculating the IoU for each of the possible pairs of melt contours that can be formed from the set of melt pool contours available at a given power and velocity. The mean and standard deviation of this IoU score is shown here. The error bar indicates $\pm$1$\sigma$ of variation.} 
\label{fig:iou_gt_depth}
\end{figure}

In order to examine the variation between successive cross-sections at the same power-velocity combination, we study the statistics of the melt pool dimensions and contour shape as a function of energy density. To examine the variation in the melt pool dimensions, we extract the depth of the melt pool for each cross-section available in the dataset. These melt pool depths are aggregated based on their associated laser power and scan velocity to compute the mean and standard deviation, as shown in Figure \ref{fig:iou_gt_depth}a). The variability of the melt pool shape is quantified by examining the overlap between the melt pool cross-sections present at a specific power-velocity combination. As the Intersection-over-Union score is defined between two sample objects, we compute the IoU score between every possible pair of $n$ melt pool cross-sections at a given power-velocity specification to obtain ${n \choose 2}$ IoU values. We report the mean and standard deviation of these IoU values in \ref{fig:iou_gt_depth}b). Notably, while the melt pool depth follows a straight-forward relationship with the laser energy density, there is much more variation in the melt pool contour shapes as a function of energy density. The mean of all ${n \choose 2}$ IoU values available for the power-velocity combinations in the dataset is 0.870.

\FloatBarrier

\section{Machine Learning Architecture Details}

The hyperparameter and architectural selections used for the Vision Transformer, Temporal Transformer, and U-Net models respectively are shown in Table \ref{table:combined}.

\begin{table}[h!]
\centering
\caption{Combined Implementation Details}
\label{table:combined}

\begin{tabular}{@{}llll@{}}
\toprule
Hyper-Parameter                  & Vision Transformer & Temporal Transformer & U-Net                            \\ \midrule
Patch/Token Dimension      & 768                & 256                   & -                                \\
Encoder Layers             & -                  & -                     & 4                                \\
Expansion Channels         & -                  & -                     & [64, 128, 256, 512, 1024]        \\
Contraction Channels       & -                  & -                     & [1024, 512, 256, 128, 64]        \\
Feed-Forward Dimension     & 3072               & 1024                  & -                                \\
Heads                      & 12                 & 4                     & -                                \\
ResNet Channel Dimensions  & -                  & [16, 32, 64, 128]     & -                                \\
Layers                     & 12                 & 4                     & -                                \\
Dropout                    & 0.1                & 0.1                   & 0.1                              \\
Learning Rate              & $1 \times 10^{-4}$ & $1 \times 10^{-4}$    & $1 \times 10^{-4}$               \\
Weight Decay               & $1 \times 10^{-3}$ & $1 \times 10^{-3}$    & $1 \times 10^{-3}$               \\
Epochs                     & 50                 & 50                    & 50                               \\
\bottomrule
\end{tabular}
\end{table}

\end{document}